\documentclass[10pt,twocolumn,letterpaper]{article}

\usepackage[pagenumbers]{cvpr} % To force page numbers, e.g. for an arXiv version

\newcommand{\inlinesection}[1]{\noindent \textbf{#1}$\,\,\,$}
\renewcommand{\paragraph}[1]{\medskip \inlinesection{#1}}
\newenvironment{smalltable}[1]{ 
    \renewcommand{\arraystretch}{1.0}
    \renewcommand{\tabcolsep}{0.7mm}
    \aboverulesep = 0.3mm % 0.605mm
    \belowrulesep = 0.5mm % 0.984mm
    \footnotesize \begin{tabular}{#1}
}{ \end{tabular} \vspace{2mm} }

\definecolor{deemph}{gray}{0.5}
\newcommand{\gc}[1]{\textcolor{deemph}{#1}}

\definecolor{cvprblue}{rgb}{0.21,0.49,0.74}
\usepackage[pagebackref,breaklinks,colorlinks,allcolors=cvprblue]{hyperref}
\usepackage[accsupp]{axessibility}

\def\paperID{****} % *** Enter the Paper ID here
\def\confName{CVPR}
\def\confYear{2025}

\title{Gaze-LLE: Gaze Target Estimation via Large-Scale Learned Encoders}

\author{Fiona Ryan$^1$ \quad Ajay Bati$^1$ \quad Sangmin Lee$^2$ \quad Daniel Bolya$^1$ \quad Judy Hoffman$^{1*}$ \quad James M. Rehg$^{3*}$ \\
$^1$Georgia Institute of Technology \quad $^2$Sungkyunkwan University\\$^3$University of Illinois Urbana-Champaign\\
{\tt\small \{fkryan,abati7,dbolya,judy\}@gatech.edu \quad sangmin.lee@skku.edu \quad jrehg@illinois.edu}
}

\begin{document}
\maketitle

{\let\thefootnote\relax\footnote{{*Equal contribution.}}}

\begin{abstract}
 We address the problem of gaze target estimation, which aims to predict where a person is looking in a scene. Predicting a person’s gaze target requires reasoning both about the person’s appearance and the contents of the scene. Prior works have developed increasingly complex, hand-crafted pipelines for gaze target estimation that carefully fuse features from separate scene encoders, head encoders, and auxiliary models for signals like depth and pose. Motivated by the success of general-purpose feature extractors on a variety of visual tasks, we propose Gaze-LLE, a novel transformer framework that streamlines gaze target estimation by leveraging features from a frozen DINOv2 encoder. We extract a single feature representation for the scene, and apply a person-specific positional prompt to decode gaze with a lightweight module. We demonstrate state-of-the-art performance across several gaze benchmarks and provide extensive analysis to validate our design choices.
 Our code and models are available at: \url{http://github.com/fkryan/gazelle}.
\end{abstract}

%%%%%%%%%% INTRO %%%%%%%%%%
\section{Introduction}
\label{sec:intro}

Gaze is an important component of human behavior, giving insight into how a person interacts with the world around them. A person's visual attention indicates intent during daily activities \cite{huang2015using, lukander2017inferring, fathi2012learning, wei2018and}, and plays a key role in social interactions \cite{emery2000eyes}. Humans can perform \textit{gaze-following}, which is the ability to assess where another person is looking. From childhood, we learn to follow the gaze of a social partner to engage in joint attention \cite{thoermer2001preverbal,frischen2007gaze, moore2014joint}. In conversations, we use gaze to infer who someone is talking to, or resolve what object they are talking about. 
Thus, the ability to estimate gaze targets is an essential building block for developing systems that understand human behavior.

\begin{figure}[tb]
  \centering
  \includegraphics[width=0.99\linewidth]{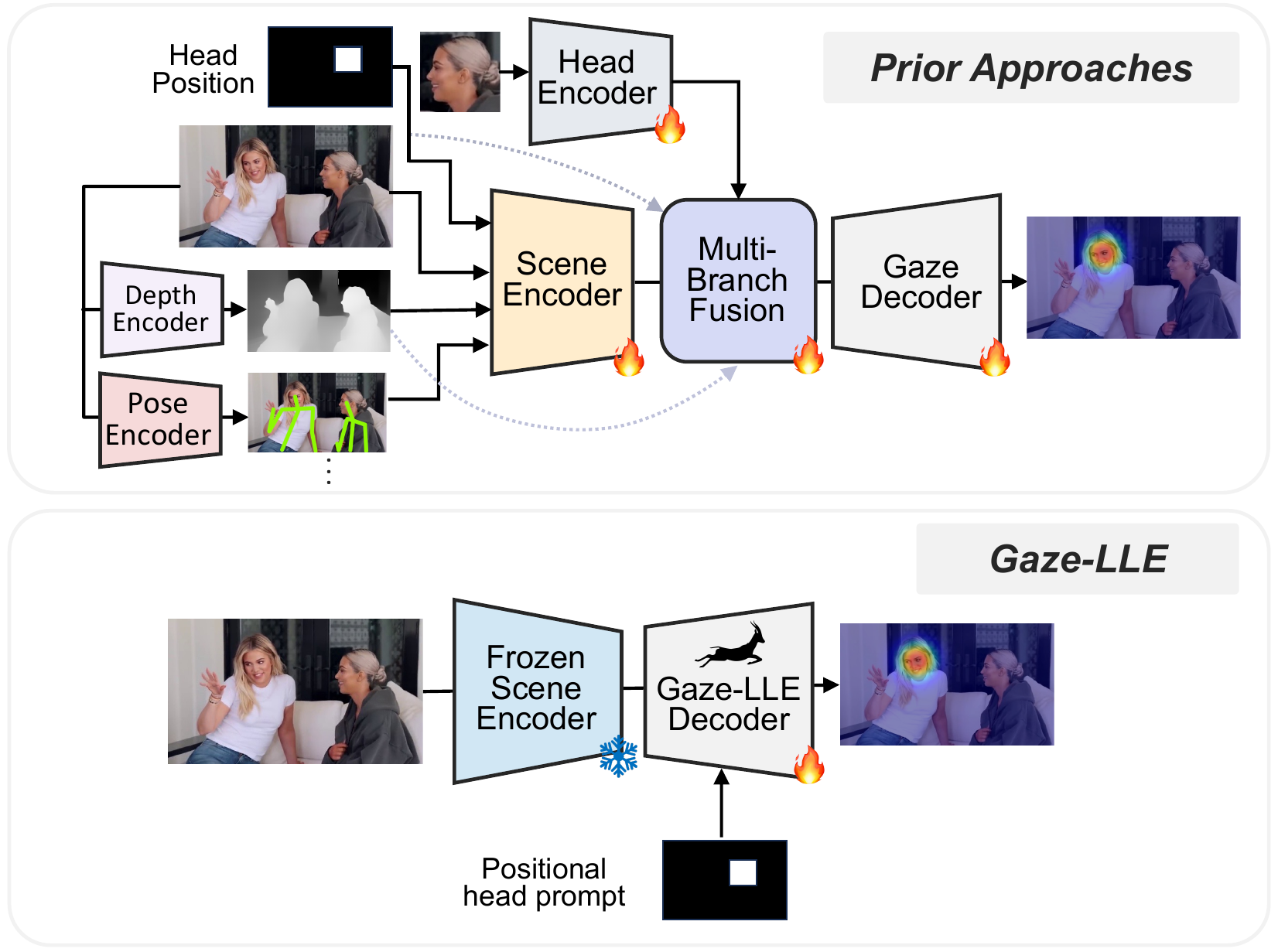}
  \caption{Prior approaches for gaze target estimation carefully fuse features from a separate head encoder, scene encoder, and auxiliary models for multimodal cues like depth and pose. We propose Gaze-LLE, a novel, streamlined approach that uses a single feature representation from a frozen image encoder and injects a person-specific positional prompt to decode gaze targets.}
  \vspace{-1em
  }
  \label{fig:teaser}
\end{figure}

A significant number of prior works have proposed specialized architectures and datasets for gaze target estimation. A key property of these architectures is a \emph{multi-branch} design, consisting of a head branch that extracts visual features from a crop of person's head and a scene branch that extracts features from the full image \cite{recasens2015they, chong2018connecting, saran2018human, lian2018believe, zhao2020learning, chong2020detecting, chen2021gaze, tafasca2024sharingan}. More recent works incorporate additional modalities such as depth \cite{fang2021dual, jin2022depth, bao2022escnet, tonini2022multimodal, tafasca2023childplay, guan2020enhanced, miao2023patch} and human pose \cite{ bao2022escnet, gupta2022modular}. While these models have achieved impressive performance, they are limited to training on  \emph{small-scale} datasets obtained by asking human annotators to label gaze targets in images. In contrast, tasks such as segmentation and depth estimation have benefited substantially from self-supervised foundation models trained on large-scale data. It is thus natural to ask: Can gaze target estimation similarly benefit from a foundation model-based approach?

In this paper, we demonstrate for the first time that pretrained visual feature representations, produced by transformer-based foundation models like DINOv2, can be leveraged via a novel architecture to yield state-of-the-art gaze estimation performance. We introduce our model for Gaze estimation via Large-scale Learned Encoders (Gaze-LLE), and show that it has two benefits. First, by establishing the feasibility of the foundation model-based approach, we enable gaze estimation to join the ranks of other dense prediction tasks in terms of leveraging and benefiting from  steady advances in the performance of foundation models. 
%%Given the difficulty and cost of reliably annotating gaze targets in images, the ability to leverage large scale data via self-supervised foundation models is a significant advance. 

Second, we show that by leveraging a powerful general purpose backbone we can simplify the model architecture significantly, reducing the learnable parameters by 1-2 orders of magnitude, and reducing the complexity of the training process while improving its efficiency. These architectural benefits are illustrated in Fig.~\ref{fig:teaser}, which contrasts Gaze-LLE against the standard multi-branch approach that uses multiple specialized encoders to capture gaze cues.
%to capture features such as depth and pose, 
%%with our novel architecture. Requiring multiple encoders and modalities for a single downstream task is undesirable in terms of ease of use, adaptability, and redundancy in scene processing per-person. Moreover, 
%%aligning features between the scene and head branches, and among the other encoders, 
This approach requires the careful fusion of different learned representations, taking their spatial relationships into account, with complex loss functions and training procedures. In contrast, Gaze-LLE leverages advances in general-purpose foundational feature representations that can solve dense prediction tasks, such as depth estimation, using only linear projection without representation tuning~\cite{oquab2023dinov2}.

%%transformer-based pretrained feature extractors such as DINOv2, which shows strong performance on dense prediction tasks like depth estimation and instance segmentation \textit{without additional training} (only learning a linear projection, see \cite{oquab2023dinov2}). Its features thus seem to \textit{already} contain information about the modalities that prior gaze estimation works need auxiliary models for, allowing us to base our approach on a single \textit{frozen} encoder.

Interestingly, merely substituting DINOv2 as a backbone in prior gaze architectures doesn't perform well. In fact, this results in significantly \textit{worse} performance than the original backbones (Tab.~\ref{tab:motivation}). Our solution is the design of our novel Gaze-LLE decoder, which adapts DINOv2 for gaze prediction. In addition, we provide substantial analysis of the challenges arising in leveraging foundation models for our task, along with extensive empirical experiments to quantify architectural decisions and tradeoffs. 

Our contributions are as follows: We introduce the novel Gaze-LLE architecture (Sec.~\ref{sec:method}) containing a specially-designed decoder that solves the problem of leveraging vision foundation models for gaze target estimation. We identify the key technical challenges and explain why the naive use of large-scale models is ineffective (Sec.~\ref{sec:motivation}), and we validate the optimality of our design decisions (Sec.~\ref{sec:analysis}).
The Gaze-LLE model is \textbf{streamlined}, with just $\sim5\%$ of the trainable parameters used in most prior methods (Tab.~\ref{tab:gazefollow_results}); \textbf{powerful}, achieving state-of-the-art performance across the three main gaze estimation benchmarks (Sec.~\ref{sec:results_main}); \textbf{general}, exhibiting strong cross-dataset performance \textit{without finetuning} (Tab.~\ref{tab:cross_dataset}); and \textbf{easy to train}, achieving state-of-the-art in $<$ 1.5 GPU-hours (Fig.~\ref{fig:convergence}). We release our code and models in the hope that even more powerful gaze estimators can be developed from Gaze-LLE.

\begin{table}[t]
    \centering
    \begin{smalltable}{llccc}
         \toprule
          Method&Encoder&  AUC $\uparrow$&  Avg L2 $\downarrow$ & Min L2 $\downarrow$ \\
          \midrule
          Chong et al. \cite{chong2020detecting} & \textbf{Original (Res50)} & \textbf{0.921} & \textbf{0.137} & \textbf{0.077}\\
          &Trained DINOv2 ViT-B  & 0.908 & 0.167 & 0.101 \\
          &Frozen DINOv2 ViT-B  & 0.875 & 0.191 & 0.125 \\
          \midrule
          Miao et al. \cite{miao2023patch} & \textbf{Original (Res50)}& \textbf{0.934} & \textbf{0.123} & \textbf{0.065} \\
          &Trained DINOv2 ViT-B  & 0.910 & 0.152 & 0.093 \\
          &Frozen DINOv2 ViT-B  & 0.892 & 0.173 & 0.109 \\
          \midrule
          Gupta et al.~\cite{gupta2022modular} & \bf Original (EfficientNet-B1) & \bf 0.933 & \bf 0.134 & \bf 0.071 \\
         $^\text{(image-only)}$&  Trained DINOv2 ViT-B & 0.912 & 0.155 & 0.090 \\
         & Frozen DINOv2 ViT-B & 0.894 & 0.184 & 0.116 \\
          \bottomrule
    \end{smalltable}
    \vspace*{-2mm}
    \caption{Existing gaze architectures 
    do not leverage features from large transformer models effectively.
    We replace the scene encoder in 3 existing open source methods with the DINOv2 ViT-B backbone and evaluate on GazeFollow (see Supp. Sec.~\ref{sec:integration} for details).
    Using DINOv2 does \textit{not} improve performance---whether or not its parameters are frozen.}
    \label{tab:motivation}
    \vspace{-1em}
\end{table}

%%%%%%%%%% RELATED WORK %%%%%%%%%%
\section{Related work}
\label{sec:relatedwork}
The dominant approach to gaze target estimation is a \emph{multi-branch fusion} approach, in which an initial encoder is followed two or more analysis branches that work in parallel to extract specific cues for gaze estimation. These branches converge in a fusion module which produces an integrated representation which is then decoded into the output heatmap, with end-to-end training of the entire pipeline. 
In contrast, the goal of this paper is to show that SotA performance can be obtained by processing the feature representation produced by a frozen foundational visual encoder, using a novel decoder architecture. We are the first to demonstrate the feasibility of using a frozen large-scale encoder for this task, and the design of our decoder architecture is novel relative to prior gaze estimation works.

The origin of the multi-branch approach is Recasens et al. \cite{recasens2015they}, which also introduced the GazeFollow dataset. Their two-branch architecture consisted of  a scene branch to estimate scene saliency, and a head branch to refine the saliency map for a specific person. This approach was adopted by many subsequent works~\cite{saran2018human, recasens2017following, chong2017detecting, lian2018believe, zhao2020learning, chong2020detecting, chen2021gaze, wang2022gatector, hu2022gaze, tafasca2024sharingan}. More recent works extended the paradigm by incorporating additional cues via auxiliary models for depth \cite{fang2021dual, jin2022depth, bao2022escnet, tonini2022multimodal, tafasca2023childplay, guan2020enhanced, miao2023patch}, body pose \cite{bao2022escnet, gupta2022modular}, 3D head direction \cite{fang2021dual, horanyi2023they}, eye location \cite{fang2021dual}, and object detections \cite{hu2022gaze}. 

\begin{figure*}[t]
  \centering
  \includegraphics[width=0.9\linewidth]{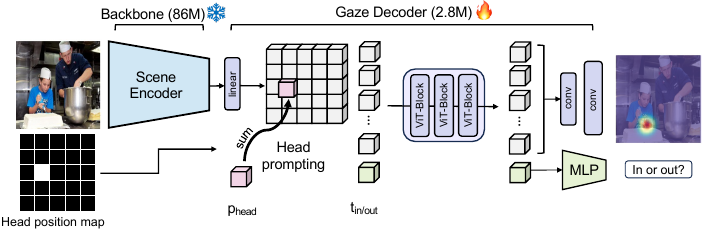}
  \caption{We introduce Gaze-LLE, a new framework for gaze estimation that learns a small gaze decoder on top of a frozen DINOv2 backbone. Using this backbone, we first extract scene tokens from an RGB image and project them to $d_{\text{model}}$ with a linear layer. We then perform \textit{head prompting} by adding a learned head position embedding $p_\text{head}$ to the scene tokens at a given person's head location. Next, we update the scene tokens and an optional learnable auxiliary in/out prediction task token $t_\text{in/out}$ with 3 transformer layers. Finally, we upsample and decode the scene tokens into a heatmap and use the in/out task token to predict if the gaze target is in or out of frame.}
  \label{fig:architecture}
\end{figure*}

A key property of gaze estimation is the need to integrate features extracted from the head region of the target person with other scene cues. This ensures that head pose, for example, is correctly interpreted in the context of the scene. Multi-branch approaches~\cite{lian2018believe, chong2020detecting, fang2021dual, bao2022escnet, jin2022depth, tonini2022multimodal, gupta2022modular, miao2023patch, horanyi2023they, tafasca2023childplay, tafasca2024sharingan} solve this problem by using the head representation as an input to the scene branch, thereby requiring the scene to be encoded separately for each person, and by carefully crafting fusion mechanisms that combine feature representations across the head branch, scene branch, and other branches. Additionally, many train with a complex multitask objective in order to supervise each encoder differently \cite{fang2021dual,jin2022depth,tonini2022multimodal,bao2022escnet,gupta2022modular, tafasca2023childplay, tafasca2024sharingan}. These complex architectures can be challenging to train and often converge slowly, as illustrated in Fig.~\ref{fig:convergence}. In contrast, we provide a head position prompt as a separate input to our unified decoder architecture (see Fig.~\ref{fig:architecture}), and all scene cues are extracted within the decoder, eliminating the need for a separate head branch and for multi-task objectives and fusion modules.

Among the multi-branch architectures, the two-branch approach of Tafasca et. al.~\cite{tafasca2024sharingan} is the most closely-related to this paper. They use a large transformer-based backbone for the scene analysis branch, which receives the head branch representation as input. While they initialize their backbone with pretrained weights, it is still trained end-to-end, with the head branch producing a specialized feature representation. In contrast, we demonstrate that 1) head analysis can also be directly integrated into the decoder, eliminating the need for a head branch and further simplifying and streamlining the architecture, and 2) frozen large-scale foundational encoders give superior performance with two orders of magnitude fewer learned parameters. Gaze-LLE produces higher accuracy on all datasets (see Tab.~\ref{fig:results}).

Some prior  works~\cite{tu2022end,tu2023joint,tonini2023object} have explored an alternative formulation of gaze target estimation as a set detection problem, where a model based on DETR \cite{carion2020end} jointly predicts the location of all heads and their accompanying gaze targets. While this avoids the need for a separate head detection step, it requires complex training, and these works use the ground truth gaze at inference time for matching - which is not consistent with practical use cases and prevents comparison with most methods (see Supp. Sec.~\ref{sec:detection}).
Other prior works have estimated 3D gaze direction from facial appearance \cite{kellnhofer2019gaze360, zhang2015appearance, fan2019understanding} without identifying the gaze target, and some early approaches estimated head orientation \cite{stiefelhagen1999modeling, yucel2009resolution} to identify gaze targets.

Another area of gaze behavior recognition involves the joint analysis of multi-person social gaze behaviors, such as shared attention (when 2 people are looking at the same gaze target) \cite{hoffman2006probabilistic, fan2018inferring, sumer2020attention, nakatani2023interaction, park2013predicting, soo2015social}, mutual gaze (when a pair of people is looking at each other) \cite{marin2011here, marin2014detecting, palmero2018automatic, marin2019laeo, doosti2021boosting, marin21pami}, and other gaze-related social structures \cite{fan2019understanding, fathi2012social, de2023temporal}. Such analysis can be used in 
assessing and understanding social behaviors for conditions like autism \cite{li2023automated, chong2020detecting, chong2017detecting}, and may benefit from an approach like ours where social context is naturally encoded within a shared scene representation.

%%%%%%%%%% METHOD %%%%%%%%%%

\section{Gaze-LLE}

\textbf{Problem Definition} Given an RGB image $x_{\text{img}} \in \mathbb{R}^{3 \times H_{in} \times W_{in}}$ and the bounding box for a particular person's head $x_{\text{bbox}} \in \mathbb{R}^4$, we predict a heatmap $\mathcal{H} \in [0, 1]^{H_{out} \ \times H_{out}}$, where each value represents the probability that the pixel is a gaze target. The VideoAttentionTarget and ChildPlay benchmarks include the additional task of predicting a value $y \in [0,1]$ that represents the probability that the given person's gaze target is inside the frame.

%%%%%%%%%% Model Architecture %%%%%%%%%%

\subsection{Model Architecture}
\label{sec:method}
Fig.~\ref{fig:architecture} illustrates our novel Gaze-LLE architecture,   consisting of a frozen, large-scale general-purpose \textit{scene encoder}  and a learned Gaze Decoder module.
Our gaze decoder performs \textit{head prompting} to condition outputs on a particular person, 
updates the feature representation with a small transformer module, and predicts a gaze heatmap and if the target is in-frame. We describe each component in  detail:

\paragraph{Scene Encoder}
A core component of our approach is leveraging strong visual features from a frozen, pretrained feature extractor $\mathcal{F}$, instead of learning a feature extractor end-to-end or using auxiliary models for signals like depth and pose.
$\mathcal{F}$ can be any visual feature extractor (see Sec.~\ref{sec:analysis}), but we primarily use DINOv2.
From $\mathcal{F}(x_\text{img})$, we obtain a lower resolution feature map of size $d_{\mathcal{F}} \times H \times W$, which we then use a linear layer to project to a smaller dimension $d_{\text{model}}$, yielding a feature map $x_\mathcal{F} \in \mathbb{R}^{d_\text{model} \times H \times W}$.

% \paragraph{Head Prompting}
\paragraph{Head Position Embedding}
A key consideration in our architecture is how to incorporate head position via head prompting. We find that incorporating head position \textit{after} the scene encoder (rather than before as in prior work) gives the best performance (see Sec.~\ref{sec:motivation} for a detailed discussion). We construct a downsampled, binarized mask $M$ of size $H \times W$ from the given head bounding box $x_{\text{bbox}}$ within the extracted scene feature map.
Using $M$, we add a learned position embedding $p_{\text{head}} \in \mathbb{R}^{d_\text{model}}$ to the scene tokens containing the head (see Sec.~\ref{sec:analysis} for alternatives).
The scene feature map $S$ is then:
\begin{equation}
    S = x_\mathcal{F} + (M * p_\text{head})
\end{equation}

\paragraph{Transformer Layers}
To update the feature representation for our task, we train a small learnable transformer module, $\mathcal{T}$, which uses self-attention to process the head-conditioned scene features. As input to $\mathcal{T}$, we flatten the feature map with the added head position $S$ into a scene token list $[s_1, s_2,...,s_{H\times W}]$. For the VideoAttentionTarget and ChildPlay benchmark settings, where the model also must classify whether the queried person's gaze is in or out of the frame, we prepend a learnable \textit{task token}, $t_{\text{in/out}}$, to the token list. Our token list is then:
\begin{equation}
[\underbrace{t_{\text{in/out}}}_{\text{task token}},\underbrace{s_1, s_2,...,s_{H \times W}}_{\text{scene tokens}}\;]
\end{equation}
Due to the spatial nature of our task, we add absolute 2d sinusoidal position embeddings \cite{dosovitskiy2020image} $P$ to the scene features before they are input to $\mathcal{T}$, \textit{i.e.}, $\mathcal{T}(S + P)$. By default, $\mathcal{T}$ consists of 3 standard transformer encoder layers \cite{vaswani2017attention}.

\paragraph{Prediction Heads}
From $\mathcal{T}(S+P)$, we obtain the updated scene features $S'$, and the updated task token, $t_{\text{in/out}}'$. We reconstruct $S'$ into a feature map of size $d_{\text{model}} \times H \times W$, and pass it to the gaze heatmap decoder $\mathcal{D_\text{hm}}$. $\mathcal{D_\text{hm}}$ consists of 2 convolutional layers to upsample the feature map to the output size $H_\text{out} \times W_\text{out}$ and produce a classification score for each pixel as being a gaze target or not. A 2-layer MLP $\mathcal{D}_\text{in/out}$ takes $t_\text{in/out}$ and outputs a classification score for if the queried person's gaze target is in or out of frame.

\paragraph{Training Objective}
We train our model using pixel-wise binary cross-entropy loss for the heatmap. Following prior work \cite{recasens2015they, chong2020detecting}, the supervisory signal is an $H_{out} \times W_{out}$ heatmap constructed by placing a 2D Gaussian distribution with $\sigma=3$ around each ground truth $(x,y)$ gaze annotation. For benchmark settings where the model must jointly predict if the gaze is in or out of frame, we use a multitask loss
\begin{equation}
\label{eq:multitask_loss}
    \mathcal{L} = \mathcal{L}_\text{hm} + \lambda \mathcal{L}_\text{in/out}
\end{equation}
where $\mathcal{L}_\text{hm}$ is pixel-wise binary cross entropy loss and $\mathcal{L}_\text{in/out}$ is binary cross entropy loss for the in/out prediction task weighted by $\lambda\in\mathbb{R}$. This loss is much simpler and easier to optimize than the complex multi-task losses employed in prior works. 
The backbone $\mathcal{F}$ is frozen during training.
Our model with a ViT-B backbone has $\sim$2.8M learnable parameters---\emph{significantly fewer than all prior works}.

%%%%% How to use Foundation Models %%%%%%

\subsection{Key Design Decisions for  Foundation Models}
\label{sec:motivation}

A key component of our approach is the use of a pretrained visual encoder (e.g.  DINOv2 \cite{oquab2023dinov2}) as a \textit{single} backbone for gaze target estimation, without any other auxiliary models. There are many possible ways to incorporate such an encoder, and in this section we systematically identify the relevant issues and explore the design space, providing empirical support for the architectural choices in Sec.~\ref{sec:method}.

Our first finding is that a straightforward substitution of DINOv2 into prior gaze architectures leads to consistently poor performance. In Tab.~\ref{tab:motivation}, we show the result of swapping the scene encoder in three open source gaze estimation methods with DINOv2 (see Supp. Sec.~\ref{sec:integration} for more results). Whether or not we finetune the DINOv2 backbone, it is outperformed by the supposedly ``weaker'' backbones in these prior works. This is not necessarily surprising, as prior gaze works~\cite{tu2022end,tu2023joint,jin2022depth} have found ResNet-50 to sometimes outperform more powerful architectures. But this finding reinforces the need for our Gaze-LLE solution. To gain further insight, we conducted a set of experiments on GazeFollow with a simple baseline: extract scene and head features with a frozen DINOv2, concatenate the results, and decode into a gaze heatmap (see Supp. Sec.~\ref{sec:design_choices} for full details). With this baseline, we quantify the impact of our three key architectural choices: integration of the head position, design of feature decoding, and use of a head branch (Tab.~\ref{tab:integration}).

\begin{table}[t]
    \begin{subfigure}[c]{\linewidth}
        \centering
        \includegraphics[width=0.7\linewidth]{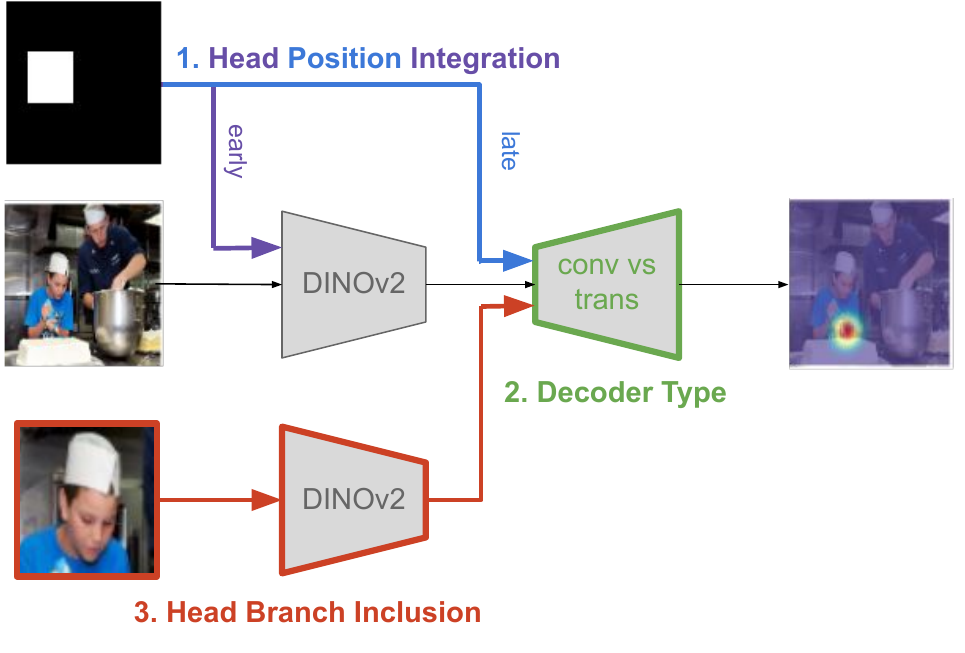}
    \end{subfigure}\vspace{0.2cm}\\
    \begin{subtable}[c]{\linewidth}
        \centering
        \begin{smalltable}{ccccccc}
            \toprule
            & (1) & (2) & (3) & \multicolumn{3}{c}{GazeFollow} \\
            & $\,$Head Integration$\,$ & $\,$Decoder$\,$ & $\,$Branches$\,$ & AUC $\uparrow$ & Avg L2 $\downarrow$ & Min L2 $\downarrow$ \\
            \midrule
            \gc{a.} & early & conv & H+S & 0.854 & 0.254 & 0.168 \\
            \gc{b.} & early & tran & H+S & 0.904 & 0.178 & 0.113 \\
            \gc{c.} & late & conv & H+S & 0.932 & 0.155 & 0.089 \\
            \gc{d.} & late & tran & H+S & \textbf{0.954} & \textbf{0.113} & \textbf{0.053} \\
            \gc{e.} & late & conv & S & 0.916 & 0.184 & 0.115 \\
            \gc{f.} & late & tran & S & \textbf{0.953} & \textbf{0.114} & \textbf{0.054} \\
        \bottomrule
        \end{smalltable}
    \end{subtable}
    \vspace*{-2mm}
    \caption{We investigate design choices across 3 axes: (1) early vs. late head integration, (2) convolutional vs. transformer decoder, and (3) using a head \& scene branch (H+S) vs. a scene branch alone (S). Row \gc{a} is the setting most similar to prior work. Conversely, we develop our final Gaze-LLE design from row \gc{f}. }
    \label{tab:integration}
\end{table}

\paragraph{Where should we inject the head position?}
The position of a person's head is an important cue in determining their gaze. Almost all prior works give head position as an extra channel to the scene branch (\textit{i.e.}, RGB + head position), which requires the scene encoder to learn how to use it when finetuning on gaze. This is problematic if we want to exploit pre-trained frozen encoders without finetuning them. We find that simply concatenating the head position channel \textit{after} extracting DINOv2 features boosts performance significantly (Tab.~\ref{tab:integration}: \gc{a} v.s. \gc{c}) compared to retraining the input projection to accept it as an additional channel.

\paragraph{How should we decode the DINOv2 features?}
Most prior works decode combined scene and head features using a stack of conv layers. This may work well when using a gaze-specialized scene encoder. However, when using a frozen DINOv2, the receptive field of a few conv layers
may be too small to extract long range gaze targets in the scene. 
We compare using a traditional 6 conv stack to decode heatmaps vs. one transformer layer into a 2-layer conv decoder.
Both arrangements have the same number of parameters, but
the transformer layer can make use of global information, thus performing better (Tab.~\ref{tab:integration}: \gc{c} v.s. \gc{d}).

\paragraph{Do we need a head branch?}
Prior works use a separate encoder that inputs a crop of the head, which is useful for understanding gaze direction. 
We hypothesize that a large-scale encoder like DINOv2 already captures gaze direction in its representation.
We compare performance with and without a head branch and find it to be nearly the same when using a transformer-based decoder (Tab.~\ref{tab:integration}: \gc{d} v.s. \gc{f}).
Notably, this doesn't occur with the conv decoder (Tab.~\ref{tab:integration}: \gc{c} v.s. \gc{e}), indicating that the relevant features are already there, but we need a transformer's global information propagation to extract them. This experiment motivated our novel head prompting design in Gaze-LLE.

%%%%%%%%%% EXPERIMENTS %%%%%%%%%%

\begin{table*}[t]
    \scriptsize
    \centering
    \begin{smalltable}{lcccccccc} 
         % & & & \multicolumn{3}{c}{| GazeFollow |} & \multicolumn{3}{c}{| VideoAttentionTarget| } \\
        \toprule
                 &&& \multicolumn{3}{c}{GazeFollow} & \multicolumn{3}{c}{VideoAttentionTarget} \\
         Method  & Learnable Params & Input & AUC $\uparrow$ & Avg L2 $\downarrow$  & Min L2 $\downarrow$ & AUC $\uparrow$ & L2 $\downarrow$  & AP{\tiny in/out} $\uparrow$ \\
         \midrule
         \textit{One Human} & & & \textit{0.924} & \textit{0.096} & \textit{0.040} &\textit{0.921} & \textit{0.051} & \textit{0.925} \\
         Recasens et al. \cite{recasens2015they} & \hphantom{$^*$}50M$^*$ & I & 0.878 & 0.19 & 0.113 & - & - & - \\
         Chong et al. \cite{chong2018connecting}  & \hphantom{$^*$}51M$^*$ &  I & 0.896&  0.187&  0.112 & 0.833 & 0.171 & 0.712 \\ 
         Lian et al. \cite{lian2018believe} & 55M &  I & 0.906&  0.145&  0.081 & - & - & - \\
         Chong et al. \cite{chong2020detecting}  & 61M &  I & 0.921&  0.137&  0.077& 0.860 &  0.134&  0.853 \\ 
         Chen et al. \cite{chen2021gaze} & \hphantom{$^*$}50M$^*$ &  I & 0.908&  0.136&  0.074 & - & - & - \\
         Fang et al. \cite{fang2021dual} & 68M &  I+D+E & 0.922&  0.124&  0.067& 0.905&  \underline{0.108} &  0.896 \\ 
         Bao et al. \cite{bao2022escnet} & \hphantom{$^*$}29M$^*$ &  I+D+P & 0.928&  0.122&  -& 0.885&  0.120 &  0.869 \\ 
         Jin et al. \cite{jin2022depth}& $>$52M$^*$ &   I+D+P & 0.920 &  0.118&  0.063 & 0.900 &  \textbf{0.104} &  0.895\\ 
         Tonini et al. \cite{tonini2022multimodal} & 92M &   I+D & 0.927& 0.141& - & \hphantom{$^\ddagger$}0.862$^\ddagger$ & 0.125 & 0.742 \\
         Hu et al. \cite{hu2022gaze} & $>$61M$^*$ & I+D+O & 0.923 & 0.128 & 0.069 & 0.880 & 0.118 & 0.881 \\
         Gupta et al. \cite{gupta2022modular}  & 35M &   I+D+P & \underline{0.943} & 0.114 & 0.056 & 0.914 & 0.110 & 0.879  \\
         Horanyi et al. \cite{horanyi2023they}$^\dagger$ & \hphantom{$^\dagger$}46M$^\dagger$  &  I+D & \hphantom{$^\dagger$}0.896$^\dagger$ & \hphantom{$^\dagger$}0.196$^\dagger$ & \hphantom{$^\dagger$}0.127$^\dagger$ & \hphantom{$^\dagger$}0.832$^\dagger$ & \hphantom{$^\dagger$}0.199$^\dagger$ & \hphantom{$^\dagger$}0.800$^\dagger$ \\
         Miao et al. \cite{miao2023patch} & 61M  &  I+D & 0.934& 0.123& 0.065 & 0.917 & \underline{0.109} & \textbf{0.908} \\
         % Horanyi et al. \cite{horanyi2023they} & $\redx$ & $\redx$& I+D & 0.932& \textbf{0.082}& \textbf{0.036} & \textbf{0.951} & \textbf{0.074} & - \\
         Tafasca et al. \cite{tafasca2023childplay} & $>$25M$^*$ &  I+D & 0.939& 0.122& 0.062 & 0.914 & \underline{0.109} & 0.834 \\
         Tafasca et al. \cite{tafasca2024sharingan} & 105M &  I & \underline{0.944} & 0.113 & 0.057 & - & \underline{0.107} & 0.891 \\
         \midrule
         \bf Gaze-LLE (ViT-B)  & 2.8M &  I & \textbf{0.956} & \underline{0.104} & \underline{0.045} & \underline{0.933} & \underline{0.107} & 0.897 \\ 
         \bf Gaze-LLE (ViT-L) & 2.9M &  I & \textbf{0.958} & \textbf{0.099} &\textbf{0.041} & \textbf{0.937} & \textbf{0.103} & \underline{0.903} \\ 
         \bottomrule
    \end{smalltable}
    \caption{Gaze target estimation results on GazeFollow and VideoAttentionTarget. We report the number of learnable parameters for each model, and if auxiliary models are used for inputs: I is image, D is depth, and P is pose, O is objects, and E is eyes. ($^*$Parameter estimate. $^\dagger$Our reimplementation, see Supp. Sec.~\ref{sec:horanyi}. $^\ddagger$Metric re-evaluated to match benchmark's calculation protocol \cite{chong2020detecting}.)}
    \label{tab:gazefollow_results} 
\end{table*}

\section{Experiments}
\label{sec:experiments}

%%\subsection{Datasets \& Metrics}
\paragraph{Datasets} We conduct experiments on GazeFollow \cite{recasens2015they} and VideoAttentionTarget \cite{chong2020detecting}, which are the primarily used benchmarks for gaze target estimation. To assess our model's generalizabity to other domains, we also include experiments on ChildPlay \cite{tafasca2023childplay}, a recent benchmark focusing on the gaze behaviors of children, and GOO-Real \cite{tomas2021goo}, which captures gaze in a retail environment.

\paragraph{Evaluation Metrics} We evaluate our model's performance by calculating heatmap AUC, which uses each heatmap pixel as a confidence score for an ROC curve, and pixel L2, which is the Euclidean distance between the argmax of the predicted heatmap and the ground truth gaze target. For GazeFollow, which contains $\sim$10 unique annotations per image, we report the distance to the average of the annotations (Avg L2), and the distance to the closest annotation (Min L2). For VideoAttentionTarget, ChildPlay, and GOO-Real, AUC is calculated by defining a tolerance region around the ground truth gaze point; for all, we follow the benchmark's specific calculation. For VideoAttentionTarget and ChildPlay, we also report the average precision (AP) for the in/out of frame prediction task.

\paragraph{Technical Details} As in prior work, our model produces a gaze heatmap of size $64 \times 64$. Because our model does not include a separate head branch that operates on a high resolution crop of each head, we use an input image size of $448 \times 448$ to capture dense details like eyes, while maintaining a small token list for computational efficiency. We conduct experiments with frozen DINOv2 ViT-B and ViT-L backbones. With DINOv2's patch size of 14, the internal feature map is size $32 \times 32$. We use an internal dimension of $d_{model}=256$ and 3 transformer layers with 8 attention heads and MLP dimension 1024. We train our model on GazeFollow for 15 epochs using the Adam optimizer, cosine scheduling with initial learning rate 1e-3, and batch size 60. We use random crop, flip, and bounding box jitter as data augmentation during training, and drop path regularization \cite{larsson2017droppath} with $p=0.1$. We finetune our GazeFollow model on VideoAttentionTarget and ChildPlay with the multitask loss in Eq.~\ref{eq:multitask_loss}. For VideoAttentionTarget, we train for 8 epochs with lr=1e-2 (in/out params) \& lr=1e-5 (other gaze decoder params), and $\lambda=1$. For ChildPlay, we train for 3 epochs with lr=2e-4 \& 1e-4 and $\lambda=0.1$.

\subsection{Main Results}
\label{sec:results_main}

\begin{figure}
    \centering
    \includegraphics[width=0.75\linewidth]{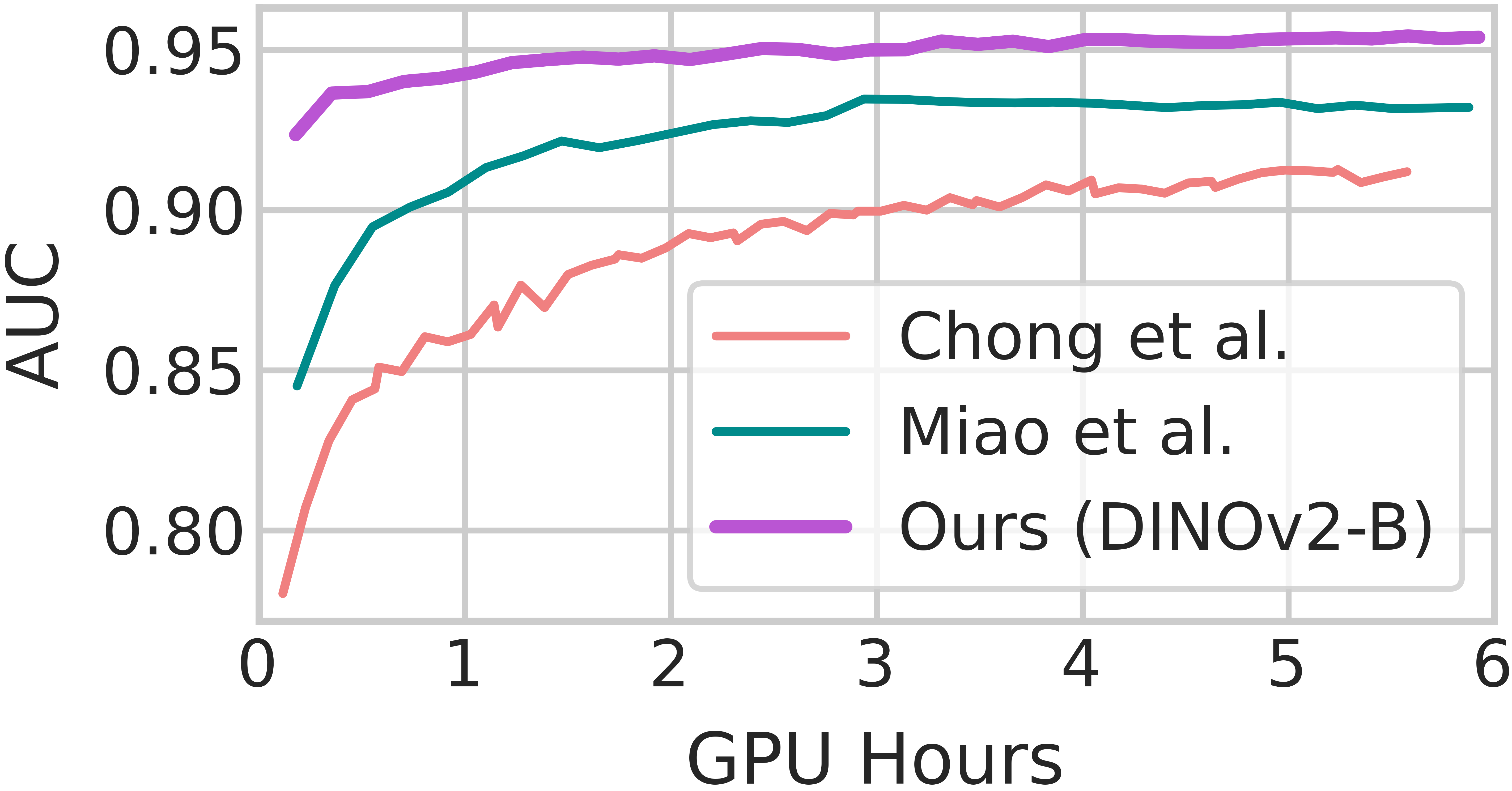}
    \vspace{-1mm}
    \caption{Training convergence: our method achieves strong results in fewer GPU hours than prior approaches.}
    \label{fig:convergence}
\end{figure}

\paragraph{Comparison to State-of-the-Art }Tab.~\ref{tab:gazefollow_results} compares the performance of Gaze-LLE with existing methods on
GazeFollow and VideoAttentionTarget. Our model achieves SotA on the AUC and L2 metrics for both datasets, while using only using a single image encoding branch and with a small fraction of the learnable parameters of prior approaches. We include results for both DINOv2 ViT-B and DINOv2 ViT-L, observing that the ViT-L backbone produces stronger results while ViT-B is still sufficient for obtaining SotA. Importantly, our ViT-B model outperforms Tafasca et al. \cite{tafasca2024sharingan}, which also uses a ViT-B backbone, but trains the backbone end-to-end along with a separate head branch and large DPT decoder. Our approach's design effectively leverages the power of the pretrained backbone, achieving stronger results with $\approx2\%$ of the learned parameters. An additional benefit of our approach is dramatically reduced training time. As illustrated in Fig.~\ref{fig:convergence}, our model converges much faster than prior methods, achieving SotA results in less than 1.5 hours on a single Nvidia RTX4090.

On VideoAttentionTarget's in/out of frame prediction task, we obtain second-best results; however our method obtains the best results when considering all metrics together. Our approach's strong, SotA results validate our hypothesis that DINOv2 features do indeed capture appropriate information for gaze target estimation, and that a single-stream design can outperform multi-stream architectures with additional modalities. To assess our model's performance on a more specialized dataset, we include results on ChildPlay in Tab.~\ref{tab:childplay}. Our ViT-B and ViT-L models both achieve SotA results across all AUC, L2, and AP. We also include the benchmark's P.Head metric \cite{tafasca2023childplay}, which assesses the precision predicting when gaze targets lie within a head bounding box; however due to its reliance on detections which are not always accurate, we find this metric tends not to correlate with the others.

 \begin{table}[h]
    \centering
    \begin{smalltable}{lcccc}
        \toprule
         Method&  AUC $\uparrow$&  L2 $\downarrow$ &  AP $\uparrow$& P.Head $\uparrow$\\
         \midrule
         Gupta et al. \cite{gupta2022modular} & 0.919 & 0.113 & 0.983 & \underline{0.694} \\
         Tafasca et al. \cite{tafasca2023childplay} & \underline{0.935} & \underline{0.107} & 0.986 & 0.663\\ 
         Tafasca et al. \cite{tafasca2024sharingan} & - & \underline{0.106} & \underline{0.990} & 0.600 \\
         \bf Gaze-LLE (ViT-B) & \textbf{0.949}  & \underline{0.106} & \bf0.994 & \bf 0.715 \\
         \bf Gaze-LLE (ViT-L) & \bf 0.951  & \bf 0.101 & \bf 0.994 & 0.662 \\
         \bottomrule
    \end{smalltable}
    \vspace*{-2mm}
    \caption{Gaze target estimation results on ChildPlay.}
    \label{tab:childplay}
\end{table}

\begin{figure*}[t]
  \includegraphics[width=0.85\linewidth]{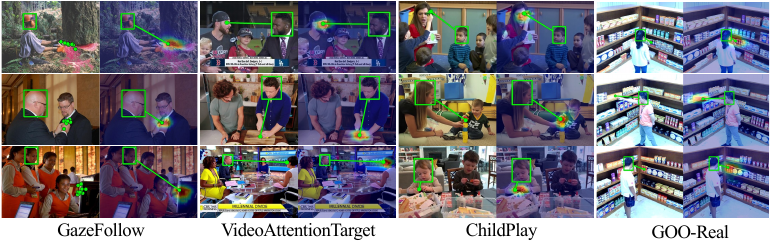}
  \centering
  \caption{Qualitative results of our GazeFollow-trained ViT-B model on GazeFollow and applied \textbf{without finetuning} to VideoAttentionTarget, ChildPlay, and GOO-Real. We show ground truth on the left and the predicted heatmap \& maximal point on the right.}
  \label{fig:results}
\end{figure*}

\begin{table}[h]
    \centering
    \begin{smalltable}{lcccccc}
        \toprule
        & \multicolumn{2}{c}{VAT} & \multicolumn{2}{c}{GOO-Real} & \multicolumn{2}{c}{ChildPlay} \\
        Method & AUC $\uparrow$ & L2 $\downarrow$ & AUC $\uparrow$ & L2 $\downarrow$ & AUC $\uparrow$ & L2 $\downarrow$ \\
        \midrule
        Chong et al.\cite{chong2020detecting}$^*$ & 0.906 & 0.119 & 0.670 & 0.334 & 0.912 & 0.121 \\
        Jin et al. \cite{jin2022depth} & 0.900 & \underline{0.104} & - & - & - & - \\
        Tonini et al. \cite{tonini2022multimodal} w/ UDA & - & - & 0.840 & 0.238 & - & - \\
        Miao et al.\cite{miao2023patch}$^*$ & 0.923 & 0.109 & 0.869 & \underline{0.202} & 0.933 & \underline{0.113} \\
        Gupta et al. \cite{gupta2022modular} & 0.907 & 0.137 & - & - & 0.923 & 0.142 \\
        Tafasca et al. \cite{tafasca2023childplay} & 0.911 & 0.123 & - & - & 0.932 & \underline{0.115} \\
        \textbf{Gaze-LLE (B)} & \underline{0.932} & \underline{0.105} & \textbf{0.901} & \textbf{0.174} & \underline{0.946} & \underline{0.114} \\
        \textbf{Gaze-LLE (L)} & \textbf{0.937} & \textbf{0.100} & \underline{0.898} & \textbf{0.175} & \textbf{0.951} & \textbf{0.101} \\
        \bottomrule
        \label{tab:zero}

    \end{smalltable}
    \vspace*{-2mm}
    \caption{Cross-dataset results on VideoAttentionTarget (VAT), GOO-Real, and ChildPlay. ($^*$Results we evaluated ourselves from the official code releases.)}
    \label{tab:cross_dataset}
    \vspace{-1em}
\end{table}

\paragraph{Cross-dataset Results} We include results of our GazeFollow-trained model applied to VideoAttentionTarget, GOO-Real, and ChildPlay \textit{without finetuning} in Tab.~\ref{tab:cross_dataset} and Fig.~\ref{fig:results}. Our approach achieves strong cross-dataset results across diverse domains and exhibits better generalization than approaches that achieve high results on GazeFollow but experience larger performance drops in cross-dataset settings (\textit{e.g.}, Gupta et al. \cite{gupta2022modular}). We attribute the strong generalizability of our method to using an encoder that is not specialized to a task or dataset, learning minimal parameters and thus not overfitting to a particular dataset, and not depending on auxiliary models, which may generalize poorly themselves. Like other methods, our model experiences the largest performance drop on GOO-Real due to (1) the large domain gap, as GOO-Real contains a unique retail environment where the user rarely faces the camera, and (2) the difference in annotation scheme - GOO-Real's ground truth is sourced from instructing participants to look at certain objects, rather than what a human annotator can reasonably infer from an image. We obtain SotA cross-dataset results, surpassing Tonini et al.'s \cite{tonini2022multimodal} method with unsupervised domain adaptation, which requires access to in-domain data at train time.

%%%%%%%%%% ANALYSIS %%%%%%%%%%
\subsection{Analysis}
\label{sec:analysis}
In this section, we provide further insight into the optimality of our design choices for Gaze-LLE. 
We investigate different backbone feature extractors and alternative strategies for head prompting and evaluate on GazeFollow.

\begin{table}[t]
    \centering
    \begin{smalltable}{lcccc}
         \toprule
         Backbone &  AUC $\uparrow$&  Avg L2 $\downarrow$& Min L2 $\downarrow$\\
         \midrule
         Supervised \cite{steiner2021train} & 0.928  & 0.151  & 0.086 \\
         % DINO \cite{caron2021emerging} & 0.935 & 0.144 & 0.079 \\
         MAE \cite{he2022masked} & 0.947  & 0.126 &  0.061 \\
         CLIP \cite{radford2021learning} & \underline{0.953}  & \underline{0.107}  & \underline{0.049} \\
         DINOv2 \cite{oquab2023dinov2} & \textbf{0.958} & \textbf{0.099} & \textbf{0.041} \\
         \bottomrule
    \end{smalltable}
    \vspace{-1mm}
    \caption{Ablation of different pretrained ViT-L backbones with Gaze-LLE on GazeFollow.}
    \label{tab:backbone_ablation}
\end{table}

\paragraph{Portability Across Backbones} While we use DINOv2 in our main experiments, Gaze-LLE can be used with any backbone. Tab.~\ref{tab:backbone_ablation} reports our model's performance with different pretrained encoders. The supervised \cite{steiner2021train} and MAE \cite{he2022masked} models are pretrained on ImageNet-1k \cite{deng2009imagenet}, while CLIP \cite{radford2021learning} and DINOv2 \cite{oquab2023dinov2} are trained on much larger data sources. Unsurprisingly, DINOv2, which is the state-of-the-art for general-purpose feature extraction on dense downstream tasks, performs best, but CLIP also achieves strong results.
As new backbones are developed, Gaze-LLE provides a framework for adapting them to gaze estimation.

\definecolor{choiceacolor}{HTML}{0070C0}
\definecolor{choicebcolor}{HTML}{C00000}
\definecolor{choiceccolor}{HTML}{00B050}

\newcommand{\choicea}[1]{\textcolor{choiceacolor}{\hphantom{(}a\hphantom{)}} #1}
\newcommand{\choiceb}[1]{\textcolor{choicebcolor}{\hphantom{(}b\hphantom{)}} #1}
\newcommand{\choicec}[1]{\textcolor{choiceccolor}{\hphantom{(}c\hphantom{)}} #1}

\begin{table}[t]
    \centering
    \begin{subfigure}[c]{\linewidth}
        \centering
        \includegraphics[width=1.0\linewidth]{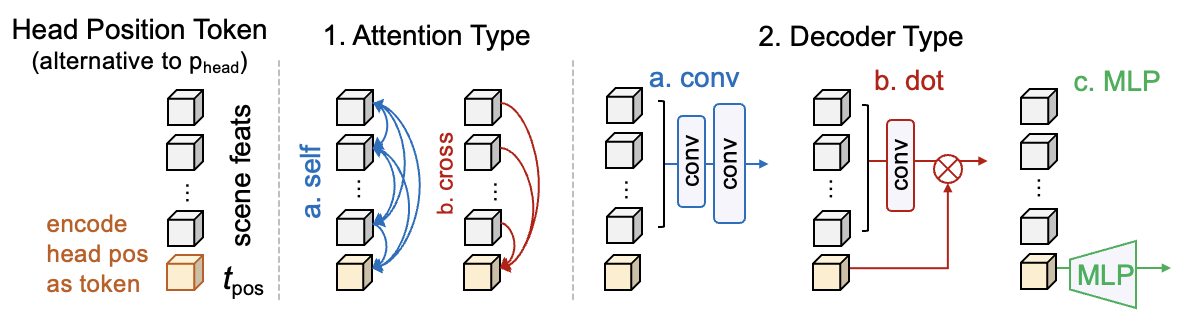}
    \end{subfigure}\vspace{0.1cm}\\
    \begin{subtable}[c]{\linewidth}
        \centering
        \renewcommand{\arraystretch}{1.03}
        \renewcommand{\tabcolsep}{3mm}
        \begin{smalltable}{lllccc}
        \toprule
        Head Prompt & (1) Attention$\,$$\,$ & (2) Decoder$\,$$\,$ & AUC $\uparrow$ & Avg L2 $\downarrow$ & Min L2 $\downarrow$ \\
        \midrule
        token & \choiceb{cross} & \choicec{mlp}
        & 0.937 & 0.117 & 0.059 \\
         & \choiceb{cross} & \choiceb{dot} & 0.945 & 0.114 & 0.055 \\
         & \choicea{self} & \choicec{mlp} & 0.939 & {0.115} & 0.058 \\
         & \choicea{self} & \choiceb{dot} & {0.952} & {0.113} & {0.052} \\
         & \choicea{self} & \choicea{conv} & \textbf{0.956} & \textbf{0.106} & \textbf{0.047} \\
        \midrule
        embedding & \choicea{self} & \choicea{conv} & \textbf{0.956} & \textbf{0.104} & \textbf{0.045} \\
        \bottomrule
        \end{smalltable}
    \end{subtable}
    \vspace*{-2mm}
    \caption{As an alternative to adding the head position embedding $p_\text{head}$ to the scene tokens, we explore representing the head's center position as an additional token, $t_\text{pos}$. We consider self attention vs. cross attention across the token list, and different ways to decode the heatmap from the scene tokens and $t_\text{pos}$.}
    \label{tab:prompt_types}
\end{table}

\paragraph{An Alternative Head Prompting Method} 
We also consider integrating the head position as its own token during attention as an alternative to our added position embedding $p_\text{head}$, inspired by works in point tracking and segmentation that represent positional queries as tokens \cite{jiang2021cotr, kirillov2023segment}. We construct a $\textit{head position token}$, $t_\text{pos}$, by sampling the position embedding $P$ at the head bounding box's center point and summing this with a learned embedding. We concatenate this token to the scene token list, $S$. To fuse positional information with the scene features, we consider two types of attention in the transformer layers: self attention across the full token list, which updates both $S$ and $t_\text{pos}$, and cross attention from the scene tokens to the position token, which updates only $t_\text{pos}$. Finally, we consider 3 methods for producing the heatmap from $S$ and $t_\text{pos}$: a 2-layer convolutional decoder on $S$ (as used in in our default method), replacing the second convolutional layer with the dot product between $t_\text{pos}$ and the scene feature map (like transformer segmentation methods \cite{cheng2021per, cheng2021mask2former, kirillov2023segment}), and directly regressing the $64\times64$ heatmap with a 2-layer MLP. We show results in Tab.~\ref{tab:prompt_types} and find that given the right settings, this position token can be made almost as effective as our default embedding approach---however, with the added benefit of potentially being able to decode multiple head positions at the same time (with each new position being another token). For our lightweight decoder, additional head locations already add negligible compute (see Supp. Sec.~\ref{sec:runtime}), but future work with heavier gaze decoders may benefit from this token design.

\begin{figure}
\centering
\includegraphics[width=0.85\linewidth]{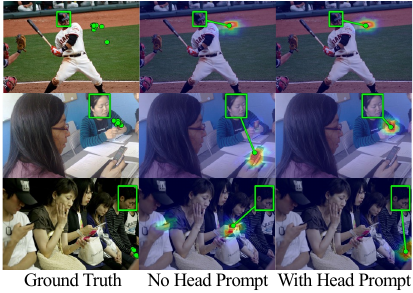}
    \caption{Without head prompting, our model succeeds on single-person cases, but cannot effectively condition gaze target estimation on the correct person in multi-person scenarios.}
    \label{fig:ablation}
\end{figure}

\begin{table}
\begin{subtable}[c]{\linewidth}
            \renewcommand{\arraystretch}{1.0}
            \renewcommand{\tabcolsep}{1.1mm}
            \centering
            \begin{smalltable}{lccc}
                \toprule
                 Prompt type &  AUC $\uparrow$&  Avg L2 $\downarrow$& Min L2 $\downarrow$\\
                 \midrule
                 No prompting & 0.926 & 0.169 & 0.105 \\
                 With prompting & \textbf{0.956} & \textbf{0.104} &  \textbf{0.045} \\
                 \bottomrule
            \end{smalltable}
            \caption{Head prompt ablation}
        \label{tab:ablation}
        \end{subtable}
        
        \begin{subtable}[c]{\linewidth}
            \centering
            \begin{smalltable}{lccc}
                \toprule
                 Prompt location &  AUC $\uparrow$&  Avg L2 $\downarrow$& Min L2 $\downarrow$\\
                 \toprule
                 Layer 3 & 0.955 & 0.108 & 0.048 \\
                 Layer 2 & 0.955 & 0.106 & 0.047 \\
                 Layer 1 (default) & \textbf{0.956} & \textbf{0.104} & \textbf{0.045} \\
                 \bottomrule
            \end{smalltable}
            \vspace*{-2mm}
            \caption{Head prompt location}
        \label{tab:prompt_layer}
        \end{subtable}
        \caption{We demonstrate the effectiveness of our head prompting mechanism (\ref{tab:ablation}), and find that injecting the head prompt before the first transformer layer in our gaze decoder module slightly outperforms later layers (\ref{tab:prompt_layer})}
        \vspace{-1em}
    
\end{table}

\paragraph{Ablating the Head Prompt} To assess the effectiveness of our head prompting mechanisms in providing the necessary information to decode a person's gaze in the absence of a head branch, we perform ablation in Tab.~\ref{tab:ablation} and Fig.~\ref{fig:ablation} by inferencing our model without providing a head prompt. The resulting performance drop shows that our head prompting strategy is effective and necessary. However, we also observe an interesting phenomenon visible in Fig.~\ref{fig:ablation}: Without a head prompt, the model still predicts a valid gaze target for at least one person in the scene - and in scenes with only one person, a head bounding box is not actually needed as input to effectively predict gaze! This result indicates that the scene representation implicitly detects heads and uses them to reason about potential gaze targets, providing further evidence for our hypothesis that a standalone head branch is not necessary. Our head prompting mechanism serves the purpose of identifying which person's gaze should be decoded in multi-person scenes.

\paragraph{Where to Perform Head Prompting} 
We also investigate \textit{where} in the decoding process to inject the head prompt. To explore the tradeoff between performance and the reduction of person-specific computation, we move the head prompting to later transformer layers in our gaze decoder, and show that this yields only small performance drops (see Tab.~\ref{tab:prompt_layer}).
This demonstrates SotA results on GazeFollow with the only per-person computation taking place in the final transformer layer and 2 convolutional layers. The ability to minimize 
person-specific computation while maintaining strong performance could provide efficient scaling for multi-person gaze analysis.
We also investigate the performance when head detections from YOLOv5 are used instead of ground truth head bounding boxes, and observe \emph{almost no degradation}. 
See Supp. Sec.~\ref{sec:yolo} for the details.

\section{Discussion}
\paragraph{Limitations} By leveraging a frozen encoder without  end-to-end training, our performance is inherently tied to the encoder quality. We find it is important to select an encoder trained on a large, diverse dataset with a dense objective (see Tab.~\ref{tab:backbone_ablation}). 
Additionally, while our method is reasonably efficient ($>$50 fps on an Nvidia RTX4090, see Supp. Sec.~\ref{sec:runtime}), the overall efficiency depends on the use of a large encoder, which may pose a challenge for embedded systems. We note that recent approaches that depend on auxiliary transformer-based depth/pose models also experience this limitation.
However, as stronger, faster general-purpose feature extractors become available, Gaze-LLE provides a way to harness them for gaze estimation.

\paragraph{Conclusion} In this work, we are the first to demonstrate that frozen foundational feature encoders can be leveraged for gaze target estimation. We propose Gaze-LLE, a new architecture that learns a gaze decoder with a novel head prompting design on top of a single, frozen DINOv2 encoder. We validate our design by achieving state-of-the-art results across four benchmarks and conducting experiments to validate the necessity and optimality of our design decisions. We hope our work opens a new chapter on gaze estimation by eliminating the need for complex multi-branch approaches via a streamlined and adaptable method that can be easily applied to new tasks and 
integrated into larger systems for understanding human behavior.

\clearpage
\paragraph{Acknowledgments} The authors thank Stefan Stojanov for helpful discussions, and the members of the Hoffman Lab and Rehg Lab for their feedback. Portions of this work were supported in part by NIH R01 MH114999, NSF \#2144194, and the NSF Graduate Research Fellowship under Grant No. DGE-2039655. Any opinion, findings, and conclusions or recommendations expressed in this material are those of the authors(s) and do not necessarily reflect the views of the National Science Foundation.

{
    \small
    \bibliographystyle{ieeenat_fullname}
    \bibliography{main}
}

\clearpage
\maketitlesupplementary

\section*{Table of Contents}
\begin{itemize}
    \itemsep0em
    \item \ref{sec:integration} Integration of DINOv2 into Existing Methods
    \item \ref{sec:design_choices} Experiment Details for Section 3.2
    \item \ref{sec:detection} Comparison to Detection Methods
    \item \ref{sec:runtime} Runtime Analysis
    \item \ref{sec:vitgaze} Comparison to ViTGaze
    \item \ref{sec:yolo} Performance with Estimated Head Bounding Boxes
    \item \ref{sec:horanyi} Reimplementation of Horanyi et al.
    \item \ref{sec:ablation} Additional Ablation Studies
    \item \ref{sec:lora} LoRA Backbones
    \item \ref{sec:viz} Additional Visualizations \& Failure Modes
\end{itemize}

\section{Integration of DINOv2 into Existing Methods}
\label{sec:integration}
In this section, we provide further details on our experiments in Tab.~\ref{tab:motivation}, which integrate DINOv2 into three existing methods: Chong et al. \cite{chong2020detecting}, Miao et al. \cite{miao2023patch}, and Gupta et al. \cite{gupta2022modular}.

\paragraph{Chong et al.} Chong et al. \cite{chong2020detecting}'s method employs separate head and scene encoders, each of which is composed of a ResNet50 + 1 additional ResNet layer. The input to the head branch is a $224\times224$ crop of the head and the input to the scene branch is the $224\times224$ scene image concatenated channel-wise with a binary map of the person's head bounding box position. The output of each encoder is a $1024\times7\times7$ feature map ($\text{channels}\times\text{height}\times\text{width}$). For our experiments, we replace the scene encoder with a ViT-Base DINOv2 encoder. Because the DINOv2 encoder produces a $768\times16\times16$ feature map, we apply average pooling with kernel size=3 and stride=2 followed by a convolutional layer with kernel size=1 and stride=1 to transform the feature map to the model's expected size of $1024\times7\times7$. We follow the rest of the existing method, which consists of an attention mechanism to re-weight the scene features based on the head features and head position, concatenation of the head and scene features, 2 convolutional encoding layers, and a 4-layer convolutional decoder. We consider 3 learning settings for the DINOv2 encoder:
\begin{enumerate}
    \item \textbf{Frozen}: We simply replace the scene encoder with the DINOv2 encoder and freeze it during training. Because the DINOv2 takes in a 3-channel RGB image, we do not concatenate the head position map to the input as in the original method.
    \item \textbf{Frozen + proj}: We alter the DINOv2 encoder's patch projection layer to take in 4 channels so that the input to the scene encoder is the concatenated RGB image and head position map like in the original method. We freeze all weights of the DINOv2 during training \textit{except} the patch projection layer.
    \item \textbf{Trained + proj}: We include the altered 4-channel patch projection layer and train the full DINOv2 encoder during training.
\end{enumerate}

\begin{table}
    \centering
    \begin{smalltable}{lcccc}
        \toprule
        DINOv2 Training & Learning rate & AUC $\uparrow$ & Avg L2 $\downarrow$ & Min L2 $\downarrow$ \\
        \midrule
        Original Method & 2.5e-4 & \textbf{0.921} & \textbf{0.137} & \textbf{0.077} \\
        \midrule
        Frozen & 2.5e-4 & 0.858 & 0.196 & 0.133 \\
        & 1.0e-4 & 0.857 & 0.201 & 0.145 \\
        & 1.0e-5 & 0.808 & 0.230 & 0.166 \\
        & 1.0e-6 & 0.726 & 0.287 & 0.218 \\
        \midrule
        Frozen + proj & 2.5e-4 & 0.875 & 0.191 & 0.125 \\
        & 1.0e-4 & 0.872 & 0.198 & 0.129 \\
        & 1.0e-5 & 0.850 & 0.212 & 0.143 \\
        & 1.0e-6 & 0.766 & 0.282 & 0.208 \\
        \midrule
        Trained + proj & 2.5e-4 & 0.876 & 0.185 & 0.120 \\
        & 1.0e-4 & \underline{0.908} & \underline{0.167} & \underline{0.101} \\
        & 1.0e-5 & 0.870 & 0.199 & 0.132 \\
        & 1.0e-6 & 0.805 & 0.260 & 0.187 \\
        \bottomrule 
    \end{smalltable}
    \caption{Comparison of integrating DINOv2 into Chong et al. \cite{chong2020detecting} with different training configurations (DINOv2 encoder learning strategy \& learning rate) on GazeFollow.}
    \label{tab:chong}
\end{table}
Tab.~\ref{tab:chong} shows our results from trying different training strategies for the DINOv2 encoder and different learning rates. We see that learning the projection layer to integrate head position as an input to the scene encoder has a significant performance gain over using the DINOv2 with RGB-only inputs, and that training the DINOv2 fully performs best. Importantly, we do not observe overfitting - the trained results are better than using the frozen DINOv2. For this method, regular training outperforms LoRA. However, all results using a DINOv2 encoder in place of the ResNet50-based scene encoder perform worse than the original method.

\paragraph{Miao et al.} Miao et al.~\cite{miao2023patch} is a more recent work that expands upon Chong et al.'s architecture by integrating estimated depth into the scene encoding and feature fusion, a global attention mechanism over the scene prior to decoding, and an additional patch-level training objective. Similar to Chong et al., Miao et al. employ head and scene encoders composed of a ResNet50 + 1 additional ResNet layer. The input to the scene branch is a 5-channel concatenation of the RGB scene image, the binary head position map, and an estimated depth map from MiDaS \cite{ranftl2020towards}. Like with Chong et al., we replace the scene encoder with a DINOv2 encoder, and use average pooling and a convolutional layer to transform the scene feature map to size $1024\times7\times7$. We consider the same training configurations as we did with Chong et al., however we change the learned patch projection to have 5 input channels to account for Miao et al.'s inclusion of depth as input. As shown in Tab.~\ref{tab:miao}, we achieve the best results by fully training the DINOv2. However, all configurations still perform worse than the original method with the ResNet50 backbone.

\begin{table}[t]
    \centering
    \begin{smalltable}{lcccc}
        \toprule
        DINOv2 Training & Learning rate & AUC $\uparrow$ & Avg L2 $\downarrow$ & Min L2 $\downarrow$ \\
        \midrule
        Original Method & 2.5e-4 & \textbf{0.934} & \textbf{0.123} & \textbf{0.065} \\
        \midrule
        Frozen & 2.5e-4 & 0.858 & 0.207 & 0.141 \\
        & 1.0e-4 & 0.859 & 0.203 & 0.138 \\
        & 1.0e-5 & 0.807 & 0.236 & 0.169 \\
        & 1.0e-6 & 0.702 & 0.297 & 0.228 \\
        \midrule
        Frozen + proj & 2.5e-4 & 0.892 & 0.173 & 0.109 \\
        & 1.0e-4 & 0.887 & 0.176 & 0.113 \\
        & 1.0e-5 & 0.859 & 0.203 & 0.137 \\
        & 1.0e-6 & 0.761 & 0.286 & 0.213 \\
        \midrule
        Trained + proj & 2.5e-4 & 0.899 & 0.165 & 0.103 \\
        & 1.0e-4 & \underline{0.910} & \underline{0.152} & \underline{0.093} \\
        & 1.0e-5 & 0.900 & 0.161 & 0.098 \\
        & 1.0e-6 & 0.847 & 0.220 & 0.149 \\
        \midrule
    \end{smalltable}
    \caption{Comparison of integrating DINOv2 into Miao et al. \cite{miao2023patch} with different training configurations (DINOv2 encoder learning strategy \& learning rate) on GazeFollow.}
    \label{tab:miao}
\end{table}

\begin{table}[t]
    \centering
    \begin{smalltable}{lcccc}
        \toprule
        DINOv2 Training & Learning rate & AUC $\uparrow$ & Avg L2 $\downarrow$ & Min L2 $\downarrow$ \\
        \midrule
        Original Method & 2.5e-4 & \textbf{0.933} & \textbf{0.134} & \textbf{0.071} \\
        \midrule
        Frozen + proj & 2.5e-4 & 0.893 & 0.180 & 0.113 \\
        & 1.0e-3 & 0.894 & 0.184 & 0.116 \\
        & 1.0e-4 & 0.897 & 0.175 & 0.108 \\
        & 1.0e-5 & 0.874 & 0.199 & 0.129 \\
        & 1.0e-6 & 0.818 & 0.228 & 0.161 \\
        \midrule
        Trained + proj & 2.5e-4 & 0.908 & 0.165 & 0.099 \\
        & 1.0e-3 & \underline{0.912} & \underline{0.155} & \underline{0.091} \\
        & 1.0e-4 & 0.911 & 0.159 & 0.095 \\
        & 1.0e-5 & 0.899 & 0.167 & 0.101 \\
        & 1.0e-6 & 0.842 & 0.219 & 0.149 \\
        \bottomrule
    \end{smalltable}
    \caption{Comparison of integrating DINOv2 into Gupta et al. \cite{gupta2022modular} (Image-only variant) with different training configurations (DINOv2 encoder learning strategy \& learning rate) on GazeFollow.}
    \label{tab:gupta}
\end{table}

\paragraph{Gupta et al.} Gupta et al.~\cite{gupta2022modular}'s approach consists of of a head-centric module, scene-centric module, and heatmap decoder. The head-centric module is a ResNet18 encoder which is supervised to predict 3D gaze from the head crop. This 3D gaze prediction is processed along with the head location into spatial gaze cone, which is passed to the scene-centric module along with the image. The scene-centric module consists of a separately trained EfficientNet encoder from different scene modalities: image, predicted depth, or predicted pose. Optionally, the encoders for the different modalities may be used together with a learned weighted attention module for fusion. As the training  process calls for separately training each modality, we consider the image-only variant for our DINOv2 integration experiments. We replace the EfficientNet-B1 image encoder with DINOv2, and add an additional learned projection layer to reduce the dimension from DINOv2's output dimension of 768 to the model's internal dimension of 64. We consider both training the full encoder and freezing the encoder (with the exception of the input projection, which must accept the extra gaze cone channel). We report performance in Tab.~\ref{tab:gupta}. Like the other methods, training the DINOv2 performs better than freezing it, but still underperforms compared to the original method.

\begin{table}
    \centering
    \begin{smalltable}{lcccc}
    \toprule
    Method & Input size & AUC & Avg L2 & Min L2 \\
    \midrule
    Chong et al. - Original & 224 & 0.921 & 0.137 & 0.077 \\
    Chong et al. - Original & 448 & 0.923 & 0.138 & 0.076 \\
    Chong et al. - Trained DINOv2 & 224 & 0.908 & 0.170 & 0.101  \\
    Chong et al. - Trained DINOv2 & 448 & 0.897 & 0.169 & 0.105 \\
    \midrule
    Miao et al. - Original & 224 & 0.934 & 0.123 & 0.065 \\
    Miao et al. - Original & 448 & 0.923 & 0.151 & 0.086 \\
    Miao et al. - Trained DINOv2 & 224 & 0.910 & 0.152 & 0.093 \\
    Miao et al. - Trained DINOv2 & 448 & 0.908 & 0.154 & 0.094 \\
    \midrule
    Gupta et al. - Original & 224 & 0.943 & 0.114 & 0.056 \\
    Gupta et al. - Original & 448 & 0.939 & 0.108 & 0.052 \\
    Gupta et al. - Trained DINOv2 & 224 & 0.912 & 0.155 & 0.091 \\
    Gupta et al. - Trained DINOv2 & 448 & 0.908 & 0.170 & 0.103 \\
    \bottomrule
    \end{smalltable}
    \caption{Effect of increasing the input scene image size for Chong et al., Miao et al., and Gupta et al.'s original methods and best variants with DINOv2. We do not observe clear gains from using a larger input size.}
    \label{tab:resolution}
    
\end{table}

\paragraph{Input Size} Because we do not include a separate head branch that operates on a higher-resolution crop of the head in our main method, we use an input size of $448\times448$ instead of $224\times224$ like these prior works. To validate that our method's gains are not only a result of the larger input to the scene encoder, we retrain Chong et al., Miao et al.'s, and Gupta et al.'s original methods as well as the best variant with a DINOv2 scene encoder with scene input size $448\times448$ in Tab.~\ref{tab:resolution}. For Gupta et al.'s original method, we use their full multimodal model. We perform average pooling on the resultant scene feature maps when necessary to reduce the spatial dimensions to the expected shape for compatibility with the rest of the model. For Chong et al.'s method, the results are largely the same between using 224 vs. 448, while for Miao et al., using 448 actually decreases performance. For Gupta et al.'s architecture, increasing the resolution to 448 results in worse AUC, which is the primary metric on GazeFollow, but achieves slight gains on the L2 metrics. We thus \textit{do not} see clear improvements from using an increased input size, illustrating that a larger scene input size is not necessary when a high-resolution head crop is already provided to the model.

\section{Experiment Details for Section 3.2}
\label{sec:design_choices}

In this section, we provide further details about our experiments in Sec.~\ref{sec:motivation} that investigate early vs. late head position integration, transformer vs. convolutional decoding, and head \& scene branch vs. scene-branch only design.

\paragraph{Scene \& Head Backbones} We use a frozen DINOv2 ViT-Base backbone for both the scene branch and the head branch. For the scene branch, we use input size $448\times448$, yielding a feature map $x_\text{scene}\in \mathbb{R}^{768\times32\times32}$. Because the head occupies a smaller portion of the full-resolution image, we use input size $224\times224$ for the head branch and upsample the resulting feature map to $x_\text{head}\in \mathbb{R}^{768\times32\times32}$ so it can be concatenated with $x_\text{scene}$. We concatenate $x_\text{scene}$ and $x_\text{head}$ channel-wise to form the combined features $x\in \mathbb{R}^{1536\times32\times32}$. For the scene-only variant, we set $x=x_\text{scene}\in\mathbb{R}^{768\times32\times32}$.

\paragraph{Head Position Integration} For ``early'' integration of the head position, we change the patch projection layer of the DINOv2 scene branch to have 4 input channels (RGB + binary head position map) instead of 3. During training, we learn this patch projection layer while keeping the rest of the DINOv2 frozen. For ``late'' integration, we do not alter or train the projection layer. Instead, we downsample the binary head position map map to size $1 \times 32 \times 32$ and concatenate it with $x$ to form $x'\in\mathbb{R}^{d\times32\times32}$ where $d=1537$ or $d=769$ depending on the inclusion of the head branch. For ``early'' integration, we do not concatenate the head position map, so  $x'=x\in\mathbb{R}^{d\times32\times32}$ where $d=1536$ (head \& scene branch) or $d=768$ (scene branch only).

\begin{table}
    \begin{subtable}[t]{1\linewidth}
        \centering
        \begin{smalltable}{c}
            \toprule
            Transformer Decoder \\
            \midrule
            Linear ($d\to256$) \\
            Trans. Layer (dim=256, heads=8, mlp\_dim=1024) \\
            ConvT($256\to256$, k=2, s=2) \\
            Conv($256\to1$, k=1, s=1) \\
            Sigmoid \\
            \bottomrule
        \end{smalltable}
    \end{subtable}
    \begin{subtable}[]{1\linewidth}
        \centering
        \begin{smalltable}{c}
            \toprule
            Conv Decoder \\
            \midrule
            Conv ($d\to768$, k=1, s=1) \\
            Conv($768\to384$, k=1, s=1) \\
            Conv($384\to192$, k=2, s=2) \\
            ConvT($192\to96$, k=2, s=2) \\
            ConvT($96\to1$, k=2, s=2) \\
            Conv($1\to1$, k=1, s=1) \\
            Sigmoid \\
            \bottomrule
        \end{smalltable}
    \end{subtable}
    \caption{Architecture details for Transformer Decoder and Convolutional Decoder for experiments in Section 3.1}
    \label{tab:3_1_arch}
\end{table}

\paragraph{Decoder}
We provide architecture details for the transformer and convolutional heatmap decoders in Tab.~\ref{tab:3_1_arch}. Each produce a $64\times64$ gaze heatmap from $x'$. The convolutional decoder is based on the network design used by Chong et al. \cite{chong2020detecting} and several subsequent methods, consisting of 6 convolutional layers (each followed by batch normalization and a ReLU activation) to progressively project the feature map to a smaller dimension while upscaling it to the output heatmap size.  The transformer decoder consists of a single transformer layer of dimension 256 followed by 2 shallow convolutional layers. Both decoders have approximately the same number of learned parameters (1.85M for the scene-branch only model with late head position integration).

\paragraph{Training Details} We train the models on GazeFollow for 15 epochs using the Adam optimizer, cosine scheduling with initial learning rate 1e-3, and batch size 60. We use the same data augmentations during training that we use in our main experiments (random crop, flip, and bounding box jitter).

\section{Comparison to Detection Methods}
\label{sec:detection}
A set of recent works formulate gaze target estimation as a set detection problem, jointly predicting a set of head bounding boxes and their corresponding gaze locations \cite{tu2022end, tonini2023object, tu2023joint}. We exclude these works from our main comparisons in Sec. ~\ref{sec:results_main}  due to differences in the evaluation setting, as these methods perform bipartite matching \textit{using the ground truth gaze targets at test time}. In this section, we provide further details about the difference in evaluation setting, and provide quantitative comparison with Tonini et al. \cite{tonini2023object} in our setting using their open source codebase.

\paragraph{Formulation} Tu et al. \cite{tu2022end} provided the first set detection formulation for joint head and gaze target detection by proposing HGTTR, a DETR \cite{carion2020end}-based transformer detection framework. Given an image $x_\text{img} \in \mathbb{R}^{3 \times H_\text{in} \times W_\text{in}}$, HGTTR predicts a fixed number of $N$ human-gaze \textit{instances}, where each instance $y$ is composed of a head bounding box prediction $y_\text{bbox} \in [0, 1]^4$, a binary classification score $y_\text{class} \in [0, 1]$ indicating the probability that the instance is indeed a head, a prediction of if the gaze is in or out of frame $y_\text{in/out} \in [0, 1]$, and a gaze heatmap $y_\text{heatmap} \in [0, 1]^{H_\text{out} \times W_\text{out}}$. Notably, the difference between this setting and the traditional problem formulation (which we follow) is that a head bounding box is not given as input. Instead, the model predicts all head bounding boxes along with their associated gaze target as output.

\paragraph{Matching Algorithm} Like DETR, HGTTR uses the Hungarian algorithm \cite{kuhn1955hungarian} to determine a one-to-one mapping between the predicted instances and the ground truth instances in order to calculate loss at train time. The optimal matching is found by considering all possible mappings $w$ between the predicted instances and ground truth instances and selecting the one that minimizes
\begin{equation}
\mathcal{L}_\text{cost} = \sum_{i=1}^N \mathcal{L}_\text{match} (y_i, \hat{y}_{w(i)})
\end{equation}
where $\mathcal{L}_\text{match} (y_i, \hat{y}_{w(i)})$ is a pairwise matching cost function between the $i$-th ground truth instance $y_i$, and the predicted instance with index $w(i)$, $\hat{y}_{w(i)}$. In HGTTR, $\mathcal{L}_\text{match}$ is defined as a weighted sum of loss functions: 

\begin{equation}
\lambda_1 \mathcal{L_\text{bbox}} + \lambda_2 \mathcal{L_\text{class}} + \lambda_3 \mathcal{L_\text{in/out}} + \lambda_4 \mathcal{L_\text{heatmap}}     
\end{equation}
where $\mathcal{L}_\text{bbox}$ is an IoU loss on the predicted bounding box, $\mathcal{L}_\text{class}$ is a binary classification loss on predicting if the instance is a head or not, $\mathcal{L}_\text{in/out}$ is a binary classification loss on predicting if the gaze is in or out of frame, and $\mathcal{L}_\text{heatmap}$ is heatmap loss between the predicted and ground truth gaze target. The model always predicts $N$ instances, with $N$ being chosen to exceed the number of ground truth instances present in each image in the dataset (all existing methods use $N=20$, which is significantly larger than the typical number of people present in a single image in GazeFollow). Because there are always less ground truth instances than predicted instances, the ground truth instance list is padded with $\emptyset$ so that it is length $N$. Predicted instances mapped to $\emptyset$ are excluded from cost and loss calculation. See Tu et al. \cite{tu2022end} and DETR (Carion et al.) \cite{carion2020end} for further details on matching.

Tonini et al. \cite{tonini2023object} expand upon this formulation, training their model to also predict all objects in the scene as an auxiliary training objective, and including depth as an input. They also add a term $y_\text{vector}$ to each instance, which is a predicted gaze vector, and use this as auxiliary supervision. Their matching cost is defined as:

\begin{equation}
\label{eq:tonini}
\lambda_1 \mathcal{L_\text{bbox}} + \lambda_2 \mathcal{L_\text{class}} + \lambda_3 \mathcal{L_\text{in/out}} + \lambda_4 \mathcal{L_\text{heatmap}} + \lambda_5    \mathcal{L_\text{vector}}
\end{equation}

For all approaches, the mapping between the ground truth and predicted instances is determined by finding the closest subset of predicted instances to the ground truth based on bounding box, class, and gaze.

\begin{table}[t]
    \centering
    \begin{smalltable}{lccc}
         \toprule
         Method &  AUC $\uparrow$&  Avg L2 $\downarrow$& Min L2 $\downarrow$\\
         \midrule
         \multicolumn{4}{c}{\textit{with ground truth gaze matching}} \\
         \midrule
         Tu et al. \cite{tu2022end} & 0.917 & 0.133 & 0.069 \\
         Tu et al. \cite{tu2023joint} & 0.928 & 0.114 & 0.057 \\
         Tonini et al. \cite{tonini2023object}  & 0.922 & 0.069 & 0.029 \\
         Tonini et al.* \cite{tonini2023object}  & 0.924 & 0.068 & 0.030 \\
         \midrule
         \multicolumn{4}{c}{\textit{no ground truth gaze matching}} \\
         \midrule
         Tonini et al.* \cite{tonini2023object} & 0.767 & 0.211 & 0.148 \\
         Ours & 0.956 & 0.104 & 0.045 \\
         \bottomrule
    \end{smalltable}
    \caption{Quantitative comparison with detection-based methods on GazeFollow. The results \textit{with ground truth gaze matching} use the ground truth gaze labels to perform bipartite matching at test time, and thus are not a direct comparison to our method and prior work. The \textit{no ground truth gaze matching} results report our method compared to Tonini et al.'s model evaluated with the altered matching cost function in Equation \ref{eq:tonini2}, which excludes ground truth gaze information. ($^*$Results we obtained ourselves by running Tonini et al.'s published code.)}
    \label{tab:detection}
\end{table}

\begin{figure*}[h]
  \includegraphics[width=1.0\linewidth]{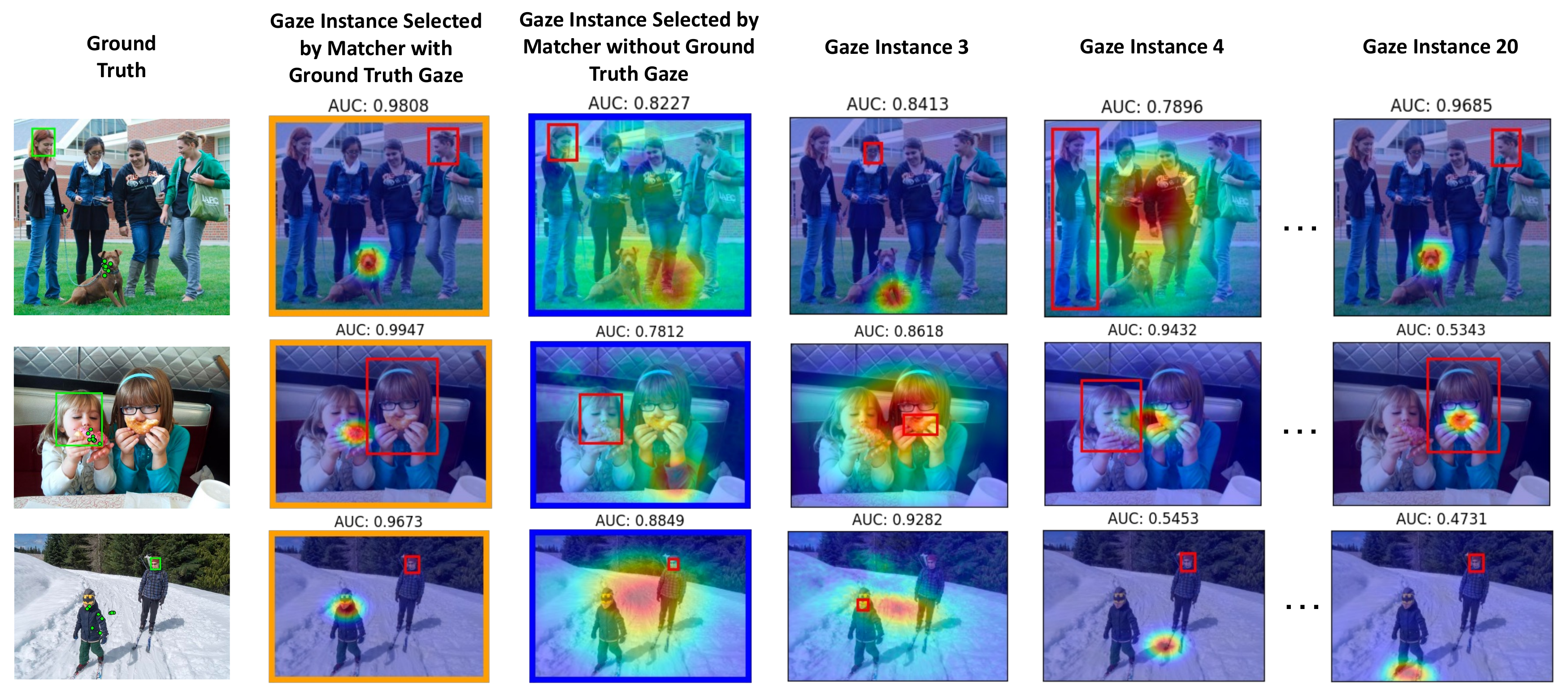}
  \centering
  \caption{We show the output gaze instances (predicted head bounding box \& gaze heatmap) from Tonini et al.'s model \cite{tonini2023object} for 3 examples. We identify the instances selected by Tonini et al.'s matching cost (which uses the ground truth gaze) and our altered matching cost (which excludes ground truth gaze and instead performs matching based on bounding box overlap). Tonini et al.'s matching algorithm selects the instance with the closest gaze prediction to the ground truth, but the bounding box prediction does not always correspond to the correct person (Rows 1-2). Additionally, we observe \textit{overdetection}, where the algorithm predicts multiple instances for the same person with different gaze heatmaps (Row 3). Without the use of ground truth gaze information, the model cannot determine which of these instances is best.}
  \label{fig:toniniMatcher}
\end{figure*}

\paragraph{Evaluation Setting} At inference time, these methods use the same matching cost $\mathcal{L}_\text{cost}$ to determine which predicted instances are evaluated against which ground truth instances, and use this to calculate the gaze performance metrics (e.g. heatmap AUC, L2 distance). This is inherently a different evaluation setting than ours because \textit{the ground truth gaze labels are used at inference time} to retrieve the predicted instances that are compared against the ground truth instances. Because the fixed number of predicted instances ($N=20$ for HGTTR) is much higher than the typical number of ground truth instances per image, a model can predict multiple instances with the same head bounding box, but different gaze targets (see Fig.~\ref{fig:toniniMatcher} for visual examples of this). In this case, the matching algorithm will match each ground truth instance to the predicted instance with the closest gaze and calculate the gaze metrics between these pairs, discarding the extra incorrect predictions from evaluation. In this way, the gaze performance metrics alone do not penalize overdetection. They characterize recall by assessing the accuracy of the predicted instances that are closest to the ground truth, but do not assess precision by penalizing the model for predicting additional instances with incorrect gaze targets. Additionally, the matching algorithm does not enforce that the ground truth instance is matched to a detection with a similar predicted head bounding box; if the heatmap loss dominates the matching cost, an instance may be selected based only on similarity between the predicted gaze heatmap and ground truth (see Fig~\ref{fig:toniniMatcher} for examples). Thus, the model does not need to correctly associate people with their respective gaze targets to achieve high performance. For these reasons, the gaze metrics in this evaluation setting are not a direct comparison against our work and prior methods that follow the traditional problem formulation.

\paragraph{Quantitative Results}
We show the reported results of the 3 detection-based methods and our results on GazeFollow in Tab.~\ref{tab:detection}. To quantitatively characterize the difference in evaluation setting, we also re-evaluate Tonini et al.'s \cite{tonini2023object} method on GazeFollow with the ground truth gaze label removed from the matching cost, using their published codebase. We alter the matching cost from Equation \ref{eq:tonini} to exclude the ground truth gaze label. This altered matching cost is defined as:
\begin{equation}
    \label{eq:tonini2}
    \mathcal{L}_\text{match}' = \lambda_1 \mathcal{L_\text{bbox}} + \lambda_2 \mathcal{L_\text{class}}
\end{equation}
With this cost, the model retrieves a prediction for each ground truth instance based only on bounding box overlap and class similarity. This reflects our use case, where the model is used to predict a gaze target for a certain person based on their head location, and does not have knowledge of the ground truth gaze. Without the use of ground truth gaze in the matching cost at inference time, we observe a significant performance drop. This quantitatively indicates the overdetection of gaze instances, as the altered matching cost results in the model selecting detections that have more bounding box overlap and class similarity\footnote{We observe that matching is mainly based on bounding box overlap. Changing the weight of class similarity in the matching cost has little effect on performance both in the original setting and our altered setting where gaze is not used in matching.} to the ground truth, but a less accurate gaze target. However, it is important to note that we do not exhaustively attempt to adapt their method to this setting (e.g. by developing a new matching algorithm for training or a non-maximal suppression method). We include this result to quantitatively demonstrate the difference in evaluation setting and use case between our method their method as-is. We note that Tu et al. \cite{tu2022end, tu2023joint} do not publish code or models so we do not re-evaluate their methods.

\paragraph{Qualitative Results}
We visualize the output instances of Tonini et al.'s default matching algorithm that uses ground truth gaze as part of the cost function, and our altered matching algorithm that does not use ground truth gaze in Fig.~\ref{fig:toniniMatcher}. The first two rows demonstrate cases where the matching algorithm chooses an instance with a predicted bounding box that is not associated with the correct person; the heatmap loss dominates the matching cost. With our altered matching function, an instance with a predicted bounding box for the correct person but incorrect gaze heatmap is retrieved. The third row shows an example of overdetection, where multiple instances are predicted with a head bounding box for the correct person, but different gaze targets. With the use of ground truth gaze during matching, the instance with the most correct heatmap is selected. However, without this ground truth information, the model does not select the best instance and produces an incorrect gaze prediction. These examples visually illustrate the difference in evaluation setting: when ground truth gaze information is used at test time, a model can achieve high performance by producing instances that capture different potential gaze targets and relying on the matching algorithm to retrieve the best instances to evaluate with. However, the gaze metrics do not characterize the model's ability to determine which of these instances are indeed gaze targets and associate them with the correct person.

\section{Runtime Analysis}
\label{sec:runtime}

\begin{figure}
    \centering
    \begin{subfigure}[]{0.8\linewidth}
        \centering
        \includegraphics[width=1.0\linewidth]{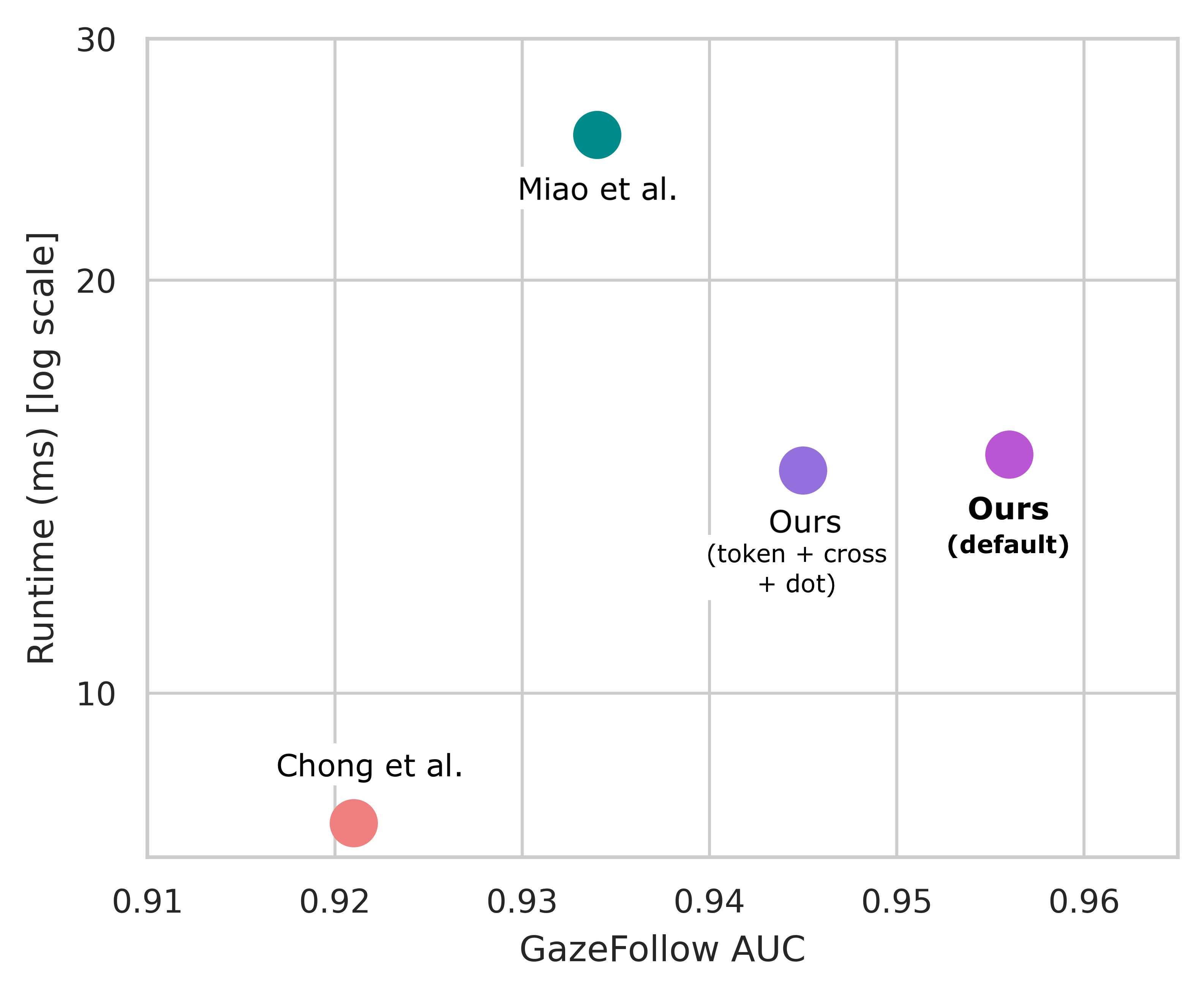}
        \caption{Runtime vs. Performance}
        \label{fig:runtime}
    \end{subfigure}
    \centering
    \begin{subfigure}[]{0.8\linewidth}
        \centering
        \includegraphics[width=1.0\linewidth]{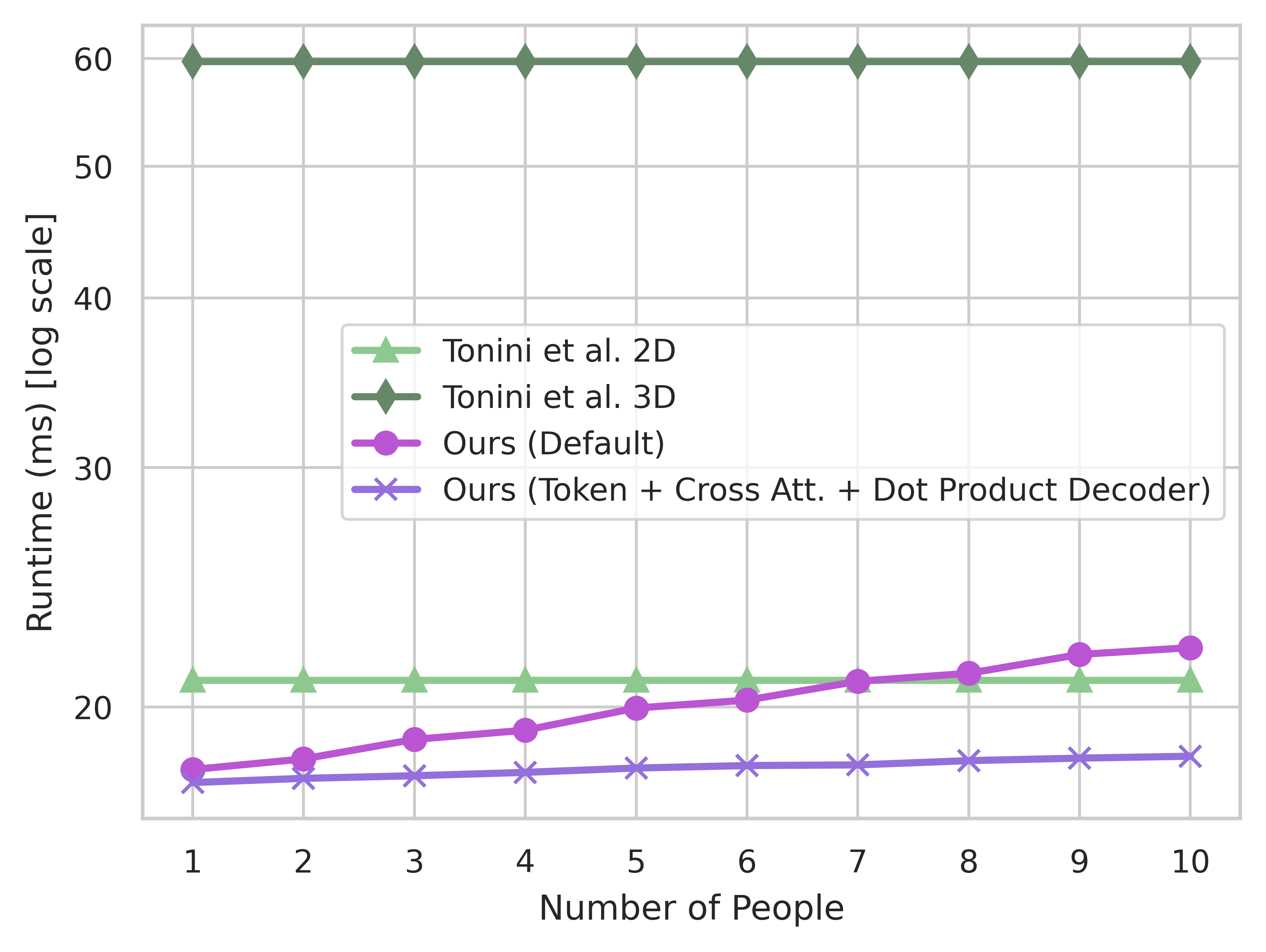}
        \caption{Runtime scaling for multi-person inference}
        \label{fig:scale}
    \end{subfigure} 
    \label{fig:run_scale}
    \caption{Runtime analysis of our approach: we show the tradeoff of inference time vs. performance (\ref{fig:runtime}), and analyze how different variants of our approach paired with a head detector scale for multi-person prediction, compared to detection methods (\ref{fig:scale}). All experiments are performed on a single NVIDIA RTX 4090 GPU.}
\end{figure}

\paragraph{Inference Speed} Our ViT-Base model runs in 15ms ($\approx66$fps) on a single NVIDIA RTX 4090 GPU. We compare the inference time of our model with existing methods in Fig~\ref{fig:runtime}. For Miao et al. \cite{miao2023patch}, we include the auxiliary depth estimation model (DPT-Hybrid\cite{ranftl2020towards}) in runtime calculation. Compared to Miao et al., our approach is both faster, and achieves better performance. In fact, the inference time of the DPT-Hybrid depth model (17ms) exceeds the entire inference time of our approach. This result highlights the benefit of using a single encoder, both in inference speed and performance. Chong et al.'s approach \cite{chong2020detecting}, which does not use any models for auxiliary modalities like depth, runs faster than our model. However, this comes with a significant drop in performance compared to our method. As shown in Tab.~\ref{tab:gazefollow_results}, recent convolutional methods all use at least one auxiliary model to augment performance. While these approaches may use faster backbones than a ViT, requiring auxiliary models ultimately increases runtime.

\paragraph{Multi-Person Scaling} We also investigate how our model's runtime scales with estimating the gaze for multiple people per image (Fig.~\ref{fig:scale}). We measure the inference time for 1-10 people per image for both our default method, and our variant that uses a head position token $(t_\text{pos})$ and decodes gaze via cross attention and a dot product with the scene features (Tab.~\ref{tab:prompt_types} configuration 1b 2b). Because the majority of our model's computation can be attributed to the DINOv2 scene encoder ($>$95\% of computation), which is run once regardless of the number of people, our model's runtime does not increase much with addition of more people (15ms for 1 person vs. 19ms for 10 people). The token variant of our model with cross attention scales even better, as it decodes gaze for all people from the same final feature map. However, as shown in Fig.~\ref{fig:runtime}, this is accompanied by a slight performance decrease.

We include Tonini et al.'s \cite{tonini2023object} detection method for comparison, which is designed to simultaneously predict the gaze and bounding boxes for all people in an image and thus has a constant runtime across different numbers of people. We include both the 2D variant (which does not use depth), and the 3D variant (which uses depth), accounting for the inference time for a DPT-Hybrid depth estimator for the 3D variant. Because our model requires head bounding boxes, we include a YOLOv5 head detector in the displayed runtimes for our model. We observe that our default method is faster than Tonini et al.'s 2D method for up to 7 people, and our token variant with cross attention is faster for all numbers of people. Due to the inclusion of running the depth model and modeling differences to include depth, the 3D version of Tonini et al.'s method is slower than the 2D version and our method. 

\section{Comparison to ViTGaze}
\label{sec:vitgaze}
We acknowledge concurrent work ViTGaze~\cite{song2024vitgaze}, which also proposes a single-branch transformer architecture for gaze target estimation based on DINOv2 pretrained weights. In contrast to Gaze-LLE, ViTGaze fully trains its ViT-S backbone (initialized from DINOv2 weights) end-to-end, and uses the attention weights between image patches as its feature representation. Gaze-LLE has the advantage of using the frozen DINOv2 features out-of-the-box, which is ideal for settings where general-purpose features are pre-computed and used for several downstream tasks. With its smaller backbone, ViTGaze is lightweight and may be better suited for on-device applications, where Gaze-LLE's ViT-B or ViT-L backbone may be too large to run. We note that ViTGaze produces predictions for the L2 metric differently than prior gaze methods: while prior work determines the maximal gaze point from a standard-sized $64\times64$ heatmap, ViTGaze uses additional postprocessing ~\cite{zhang2020distribution} to bypass the limitations of the low resolution of the output heatmap. ViTGaze also uses a higher input resolution (512).

\section{Performance with Estimated Head Bounding Boxes}
\label{sec:yolo}

\begin{table}
    \centering
    \begin{smalltable}{lccc}
        \toprule
        Method & AUC $\uparrow$ & Avg L2 $\downarrow$ & Min L2 $\downarrow$ \\
        \midrule
        ViT-B + GT & 0.956 & 0.104 & 0.045 \\
        ViT-B + YOLO & 0.955 & 0.106 & 0.047 \\
        \midrule
        ViT-L + GT & 0.958 & 0.099 & 0.041 \\
        ViT-L + YOLO & 0.958 & 0.101 & 0.043 \\ 
    \bottomrule
    \end{smalltable}
    \caption{Gaze-LLE achieves consistent results when using head detections from an out-of-the-box YOLOv5 detector instead of head ground truth bounding boxes.}
    \label{tab:yolo}
\end{table}

Tu et al. \cite{tu2022end} report that 2-stream methods suffer major performance drops when using head bounding boxes from a detector rather than the dataset ground truth. In contrast, we observe almost no performance degradation when pairing our method with a YOLOv5 head detector trained on CrowdHuman \cite{shao2018crowdhuman, yolov5crowdhuman} (Tab. \ref{tab:yolo}). This result demonstrates that our single-stream design, which uses a coarse, downsampled head position map, is less dependent on an exact head crop, and works well with out of the box head detections. Given DINOv2's strong performance on tasks such as semantic segmentation with linear probing, future work may explore integrating head detection directly into the pipeline by predicting heads from the same frozen DINOv2 features.

\section{Reimplementation of Horanyi et al.}
\label{sec:horanyi}

\begin{figure*}[]
    \begin{subfigure}[c]{0.6\linewidth}
        \centering
        \includegraphics[width=0.8\linewidth]{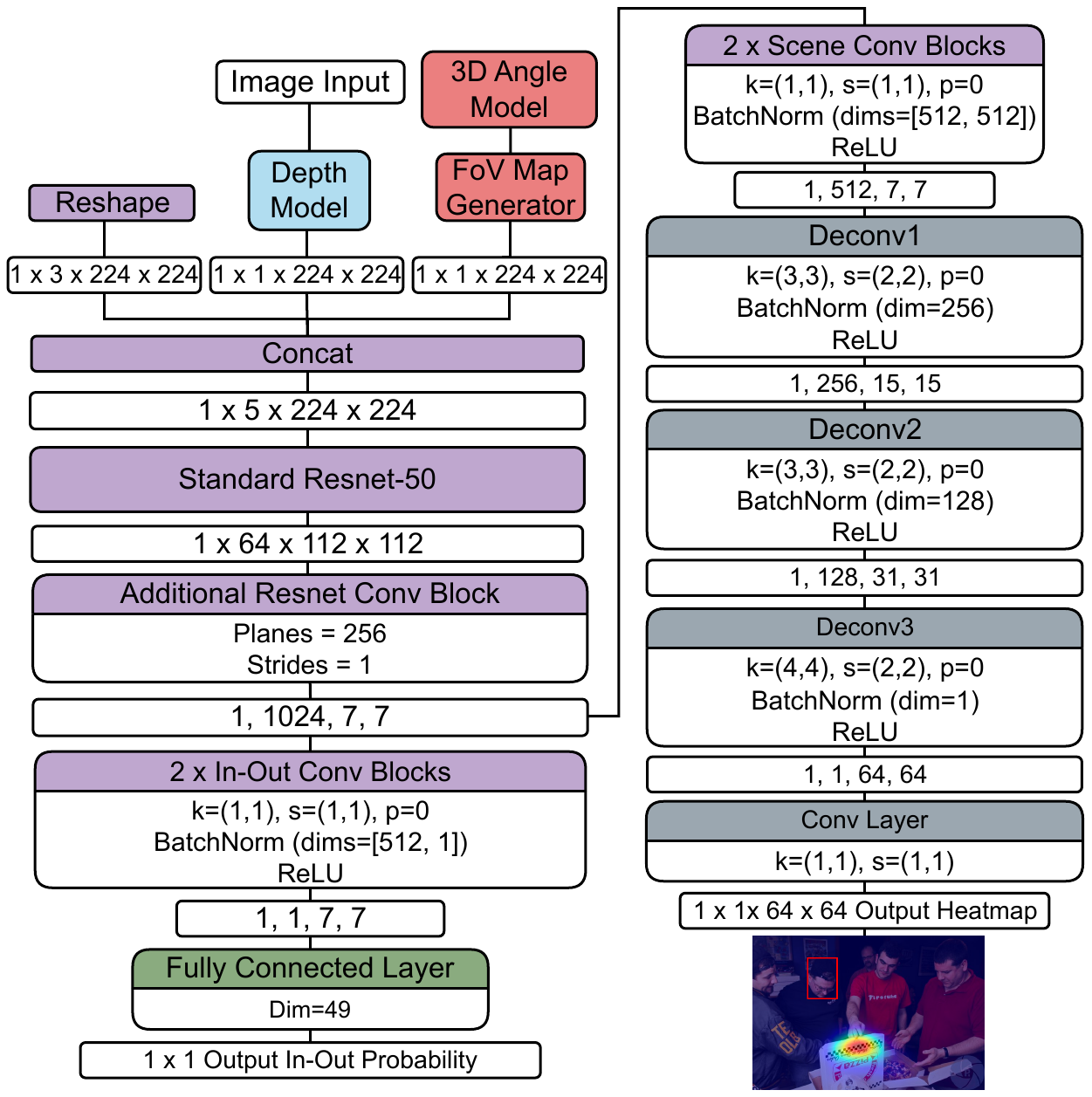}
    \end{subfigure}%
    \hfill
  \begin{subfigure}[c]{0.35\linewidth}
        \includegraphics[width=0.8\linewidth]{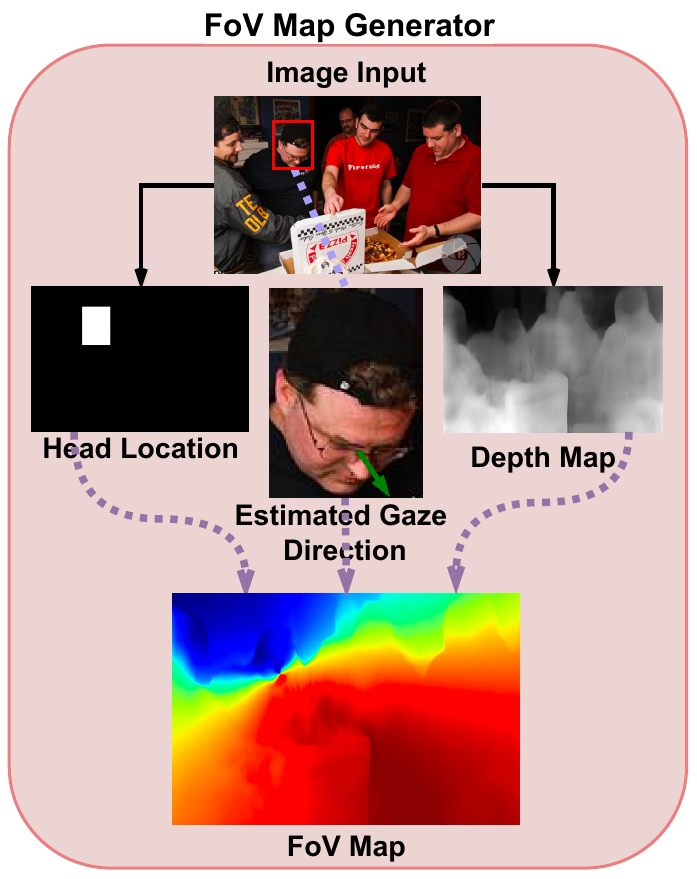}
    \end{subfigure}%
    \caption{Architecture details for our reimplementation of Horanyi et al.'s model \cite{horanyi2023they}. The model consists of a FoV Map Generator (shown on right), which uses an auxiliary 3D gaze angle estimator and an auxiliary depth model to produce an FoV map for a given person. The FoV map, estimated depth, and image are passed to a ResNet50-based encoder and convolutional decoder to produce a gaze prediction. In our experiments, we consider both freezing vs. training the 3D gaze angle estimator as part of the model.}
    \label{fig:horanyi}
  
\end{figure*}

We use our own implementation of Horanyi et al. \cite{horanyi2023they} for our main comparison. We choose to reimplement this method because the reported results are outliers among other methods, and there is imbalance between the reported metrics (\textit{e.g.}, 0.932 AUC on GazeFollow, but very low L2 error). Since the method is largely constructed from elements that are present in other works (\textit{e.g.}, constructing a ``gaze cone'' from estimated 3D gaze and depth \cite{fang2021dual, gupta2022modular, tonini2023object}, providing estimated depth as input to the scene encoder \cite{fang2021dual, gupta2022modular}, and using a ResNet50-based scene encoder + 4-layer convolutional decoder \cite{chong2020detecting, miao2023patch}), it is difficult to identify the source of large reported performance gains. There is not published code for this work. The original paper provides limited implementation details, so we follow some choices from Chong et al.'s codebase \cite{chong2020detecting}.

\begin{table}[t]
    \scriptsize
    \centering
    \begin{smalltable}{lcccccc} 
        \toprule
                & \multicolumn{3}{c}{GazeFollow} & \multicolumn{3}{c}{VideoAttentionTarget} \\
         Experiment & AUC $\uparrow$ & Avg L2 $\downarrow$  & Min L2 $\downarrow$ & AUC $\uparrow$ & L2 $\downarrow$  & AP{\tiny in/out} $\uparrow$ \\
         \midrule
         Frozen Aux. Angle & 0.869 & 0.217 & 0.146 & 0.802 & 0.234 & 0.720 \\
         Trained Aux. Angle & \textbf{0.896} & \textbf{0.196} & \textbf{0.127} & \textbf{0.832} & \textbf{0.199} & \textbf{0.800} \\
         \bottomrule
    \end{smalltable}
    \caption{Experimental results for our implementation of Horanyi et al.\cite{horanyi2023they} on GazeFollow and VideoAttentionTarget. We consider the setting where we freeze the auxiliary 3D gaze angle model vs. where we train it along with the rest of the network. }
    \label{tab:horanyi} 
\end{table}

The model consists of a 3D Field-of-View (FoV) map construction module, a scene encoder, and a gaze decoder. The FoV module uses an auxiliary depth estimator and 3D gaze angle estimator to produce an FoV heatmap for a person over the scene. The estimated depth, FoV map, and $224\times224$ map are passed to a ResNet50-based scene encoder and decoded into gaze predictions. Figure~\ref{fig:horanyi} illustrates the architecture details for our reimplementation. We use the same auxiliary models used in the original approach \cite{horanyi2023they}: Gaze360 \cite{kellnhofer2019gaze360} and Monodepth2 \cite{godard2019digging}. We follow the version of their scene encoder without non-local (NL) blocks. The FoV module uses the construction equation from Horanyi et al. \cite{horanyi2023they}:

\begin{equation}
    \footnotesize
    M_{ind} = \text{min\_max\_scaler} \left( \frac{(i-h_x,j-h_y,k-h_z)\cdot(g_x,g_y,g_z)}{\lVert i-h_x,j-h_y,k-h_z \rVert^2 \cdot \lVert g_x,g_y,g_z \rVert} \right)
\end{equation}

We make the assumption that $k$-coordinate comes from the normalization of the estimated depth map. We follow the high-level architecture described in the text: a ResNet50 trainable scene encoder, two convolutions for encoding, and a 4-layer convolutional decoder. Because details such as hidden dimensions and kernel sizes are not specified, we generally follow Chong et al.'s open-source code \cite{chong2020detecting} since Horanyi et al.'s described architecture mostly matches Chong et al.'s. We conduct experiments in two settings on the Gazefollow and VideoAttentionTarget datasets. The first setting keeps both the auxiliary gaze angle and depth estimation models frozen, as suggested in the text \cite{horanyi2023they}. In the second setting, we train the gaze angle model. For the GazeFollow experiments, we use batch size 128, learning rate 4e-4, and the Adam optimizer. For VideoAttentionTarget, we finetune the GazeFollow-trained model with batch size 32 and learning rate 1e-4. The results of these experiments are shown in Tab.~\ref{tab:horanyi}. We observe training the auxiliary gaze angle model performs better, so we report these results in the main paper.

\section{Additional Ablation Studies}
\label{sec:ablation}

\begin{table}[t]
    \begin{subtable}[c]{\linewidth}
    \centering
        \begin{smalltable}{lcccc}
            \toprule
            $d_\text{model}$ & Params & AUC $\uparrow$ & Avg L2 $\downarrow$ & Min L2 $\downarrow$ \\
            \midrule
            128 & 1.2M & \bf 0.956 & 0.106 & 0.046 \\
            256 (default) & 2.8M & \bf 0.956 & \bf 0.104 & \bf 0.045 \\
            384 & 5.0M & \bf 0.956 & 0.105 & 0.046 \\
            512 & 7.7M & 0.953 & 0.108 & 0.049 \\
            768 & 14.8M & 0.953 & 0.108 & 0.049 \\
            \bottomrule 
        \end{smalltable}
        \caption{Dimension of gaze estimation module.}
        \label{tab:dimension}
    \end{subtable}
    \begin{subtable}[c]{\linewidth}
    \centering
        \begin{smalltable}{lcccc}
            \toprule
            Layers & Params & AUC $\uparrow$ & Avg L2 $\downarrow$ & Min L2 $\downarrow$ \\
            \midrule
            1 layer & 1.2M & 0.953 & 0.115 & 0.054 \\
            2 layers & 2.0M & 0.955 & 0.108 & 0.049 \\
            3 layers & 2.8M & \bf 0.956 & 0.104 & \bf 0.045 \\
            4 layers & 3.6M & \bf 0.956 & \bf 0.103 & \bf 0.045 \\
            5 layers & 4.4M & \bf 0.956 & 0.104 & \bf 0.045 \\
            \bottomrule 
        \end{smalltable}
        \caption{Number of transformer layers.}
        \label{tab:layers}
    \end{subtable}
    \caption{We investigate the effect of different internal model dimensions and number of transformer layers for our gaze estimation module with a ViT-Base DINOv2 backbone. We observe diminishing returns as we increase the dimension and number of layers. We select $d_\text{model}=256$ with 3 transformer layers as our default configuration.}
    \label{tab:ablation}
\end{table}

We provide additional ablations for our ViT-Base model on GazeFollow in Tab.~\ref{tab:ablation}. We find there is benefit to using a smaller internal dimension for our gaze estimation module, both in performance and reduction of learnable parameters (Tab. \ref{tab:dimension}). Our model produces competitive results with prior work using only 1 transformer layer (Tab. \ref{tab:layers}); however, we achieve sizeable performance gains by increasing the number of layers to 3. Beyond 3 layers, the performance is largely stable. To balance performance with reducing learnable parameters, we select dimension 256 with 3 layers as our default configuration.

\begin{figure*}[!h]
  \centering
  \begin{subfigure}{\linewidth}
    \begin{center}
        \includegraphics[width=0.93\linewidth]{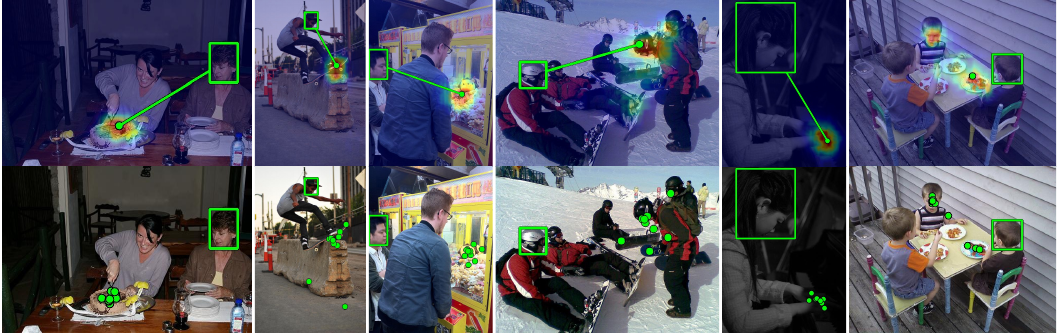}
    \caption{GazeFollow}
    \label{fig:bigviz-a}
    \end{center}
  \end{subfigure}
  \begin{subfigure}{\linewidth}
    \begin{center}
        \includegraphics[width=0.93\linewidth]{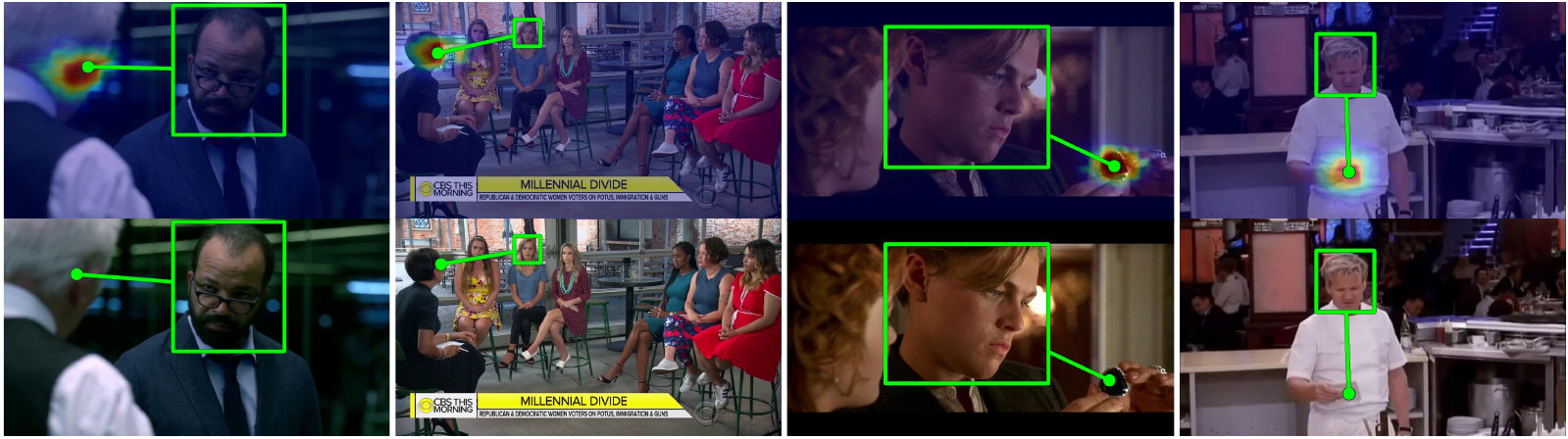}
    \caption{VideoAttentionTarget}
    \label{fig:bigviz-b}
    \end{center}
  \end{subfigure}
  \begin{subfigure}{\linewidth}
    \begin{center}
        \includegraphics[width=0.93\linewidth]{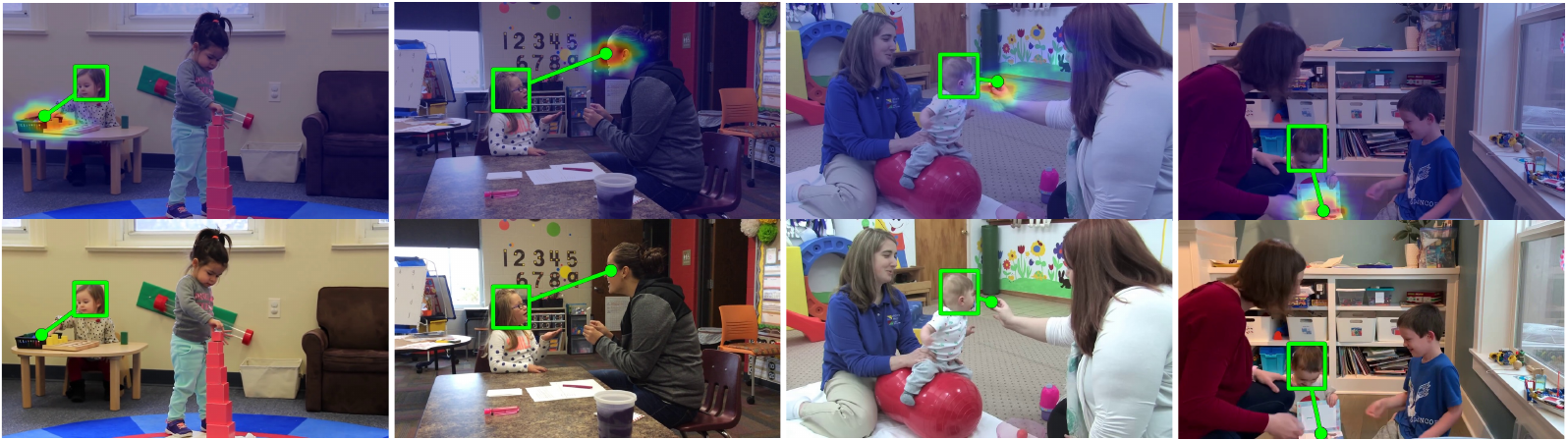}
    \caption{ChildPlay}
    \label{fig:bigviz-c}
    \end{center}
  \end{subfigure}
  \begin{subfigure}{\linewidth}
    \begin{center}
        \includegraphics[width=0.93\linewidth]{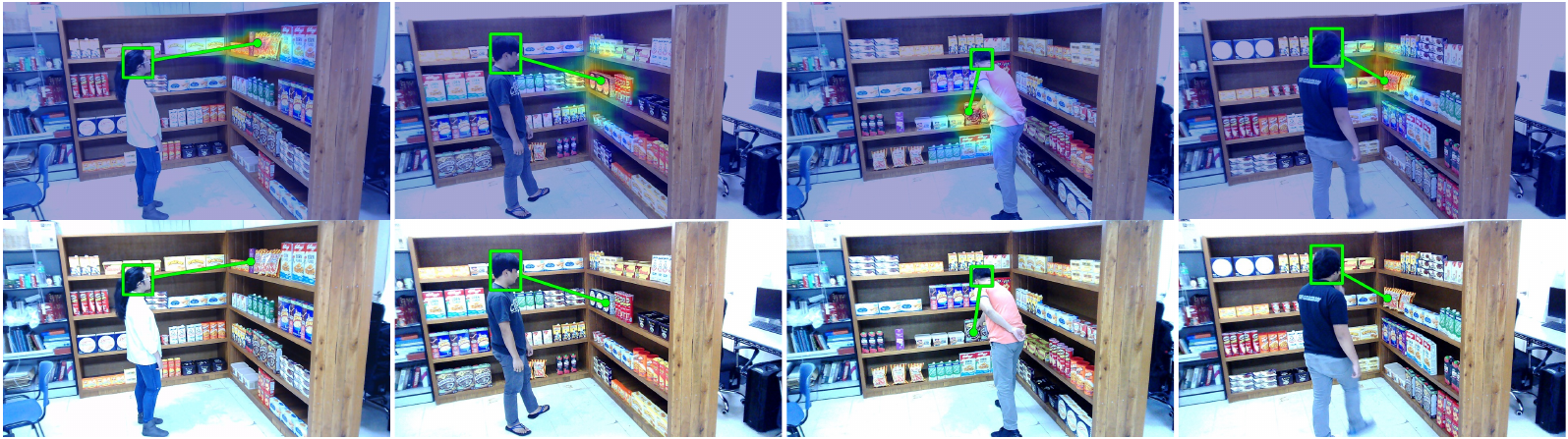}
    \caption{GOO-Real}
    \label{fig:bigviz-d}
    \end{center}
  \end{subfigure}
  \caption{Additional qualitative results on the 4 evaluation datasets: For each example, we show our model's predicted heatmap with its maximum point on the top, and the ground truth gaze annotations on the bottom.}
  \label{fig:bigviz}
\end{figure*}

\section{LoRA Backbones}
\label{sec:lora}

\begin{table}
    \centering
    \begin{smalltable}{lcccc}
         \toprule
         Backbone & Params & AUC $\uparrow$&  Avg L2 $\downarrow$& Min L2 $\downarrow$\\
         \midrule
         \textit{One Human}  & & \textit{0.924} & \textit{0.096} & \textit{0.040} \\
         \midrule
         ViT-B &  2.8M & 0.956  & 0.104  & 0.045 \\
         ViT-B + LoRA &  3.1M & 0.957  & 0.103  & 0.045 \\
         ViT-L & 2.9M & 0.958 & 0.099  & 0.041 \\
         ViT-L + LoRA & 3.7M & 0.960  & 0.097  & 0.040 \\
         \bottomrule
    \end{smalltable}
    \caption{LoRA-tuned DINOv2 Backbones}
    \label{tab:lora}
    \caption{Frozen vs. LoRA-tuned DINOv2 backboneswith Gaze-LLE on GazeFollow.}
    \label{tab:lora}
\end{table}

To investigate if training the backbone improves performance, we explore using Low Rank Adaptation (LoRA) \cite{hu2021lora} on GazeFollow in Tab.~\ref{tab:lora}. LoRA updates the backbone while introducing limited additional learnable parameters by learning weight update matrices as low rank decompositions. We update the query and value projections of the DINOv2 backbone using rank 16. We observe limited improvements, which we attribute to (1) the effectiveness of the frozen encoder's feature representation for our task and (2) that our models with frozen encoders already achieve extremely close performance to the inter-rater performance of the human annotators, which serves as a soft upper bound on the L2 metrics.

\begin{figure*}[t]
  \centering
   \includegraphics[width=1.0\linewidth]{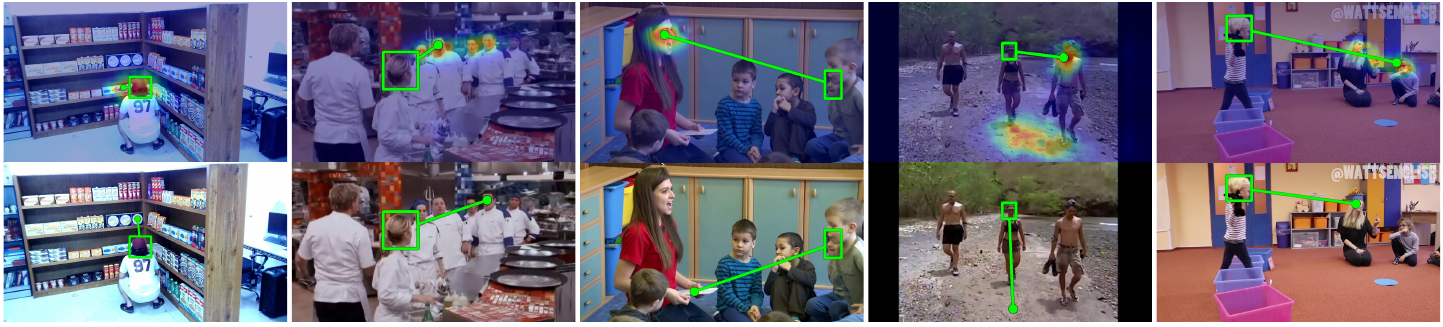}
   \caption{Lower performing cases: we observe errors in some cases where the head is facing away from the camera (examples 1-2), the head is occluded (examples 3), or the face is blurred (examples 4-5).}
   \label{fig:failures}
\end{figure*}

\section{Additional Visualizations \& Failure Modes}
\label{sec:viz}

We provide additional visualizations of our ViT-B model's predicted heatmaps on the GazeFollow, VideoAttentionTarget, ChildPlay, and GOO-Real datasets in Figure \ref{fig:bigviz}. We show examples where our model does not perform as well in Figure \ref{fig:failures}. These cases are representative of error modes we observe across the evaluation datasets. Our model is more likely to exhibit errors when the person is positioned with the back of their head  towards the camera (examples 1-2) or their face is heavily occluded (example 3). In these cases, we observe our model selects potential targets (such as faces) that are broadly in person's field of view, but does not always refine this prediction to the ground truth gaze target. It is not surprising that the model does not perform as well on these cases, as the ground truth is often inherently more ambiguous in such examples. We observe similar errors in cases where the person's face and eyes are blurred (examples 4-5), which is more common in video datasets like VideoAttentionTarget and ChildPlay. Future work may explore using temporal information from surrounding frames to resolve ambiguities in these cases.

\end{document}